\documentclass[10pt,journal,compsoc]{IEEEtran}
\ifCLASSOPTIONcompsoc
  \usepackage[nocompress]{cite}
  \usepackage{CJK}
\usepackage{amsmath}
 \usepackage{mathrsfs}
 \usepackage{algorithm}
\usepackage{algorithmic}
\usepackage{multirow}
\usepackage{amsmath}
\usepackage{amssymb}
\usepackage{xcolor}
\usepackage{setspace}
\usepackage{indentfirst}
\usepackage{epsfig}
\usepackage{graphicx}
\usepackage{subfigure}
\usepackage{rotating}
\usepackage{float}
\usepackage{stfloats}
\usepackage{verbatim}
\else
  \usepackage{cite}

\fi
\ifCLASSINFOpdf
\else
\fi
\begin{document}
%
\title{Multi-target Unsupervised Domain Adaptation without Exactly Shared Categories}
%
%
%
%

\author{Huanhuan Yu,
        Menglei Hu
        and Songcan Chen

\IEEEcompsocitemizethanks{\IEEEcompsocthanksitem The authors are with College of Computer Science and Technology Nanjing University of Aeronautics and Astronautics, Nanjing China, 211106. Corresponding author: Songcan Chen\protect\\
E-mail: \{h.yu, s.chen\}@nuaa.edu.cn.}
\thanks{}}

%
%

\markboth{}%
{Shell \MakeLowercase{\textit{et al.}}: Bare Demo of IEEEtran.cls for Computer Society Journals}
%



\IEEEtitleabstractindextext{%
\begin{abstract}
Unsupervised domain adaptation (UDA) aims to learn the unlabeled target domain by transferring the knowledge of the labeled source domain. To date, most of the existing works focus on the scenario of one source domain and one target domain (1S1T), and just a few works concern the scenario of multiple source domains and one target domain (mS1T). While, to the best of our knowledge, almost no work concerns the scenario of one source domain and multiple target domains (1SmT), in which these unlabeled target domains may not necessarily share the same categories, therefore, contrasting to mS1T, 1SmT is more challenging. Accordingly, for such a new UDA scenario, we propose a UDA framework through the model parameter adaptation (PA-1SmT). A key ingredient of PA-1SmT is to transfer knowledge through adaptive learning of a common model parameter dictionary, which is completely different from existing popular methods for UDA, such as subspace alignment, distribution matching etc., and can also be directly used for DA of privacy protection due to the fact that the knowledge is transferred just via the model parameters rather than data itself. Finally, our experimental results on three domain adaptation benchmark datasets demonstrate the superiority of our framework.
\end{abstract}

\begin{IEEEkeywords}
Unsupervised domain adaptation, model parameter adaptation, multiple target domains, knowledge transfer
\end{IEEEkeywords}}

\maketitle

\IEEEdisplaynontitleabstractindextext

%
\IEEEpeerreviewmaketitle

\IEEEraisesectionheading{\section{Introduction}\label{sec:introduction}}

%
%
%
%
\IEEEPARstart{I}{n} most traditional machine learning problems, models trained from the training data directly use for the testing data. However, in real world applications, due to different factors, such as viewpoint variation, pattern difference etc., the training and testing data are related but have different distributions. Such a distribution discrepancy results in the learned model not being well adapted to the testing data. In order to address this problem, we can transfer the useful knowledge from the related training domain and apply to the current testing domain, which is called domain adaptation (DA) [1]. In DA, the training and testing domains are respectively named as the source and target domains. In addition, since manually labeled target data could be expensive, the semi-supervised domain adaptation and the unsupervised domain adaptation (UDA) basically become a mainstream. For such two scenarios, the labels are available in the source domain, and in the target domain, the former has only a few labels, while the latter has no labels [2]. Thus, UDA is harder to find the connections between the source and target domains. In this paper, we focus on this more difficult unsupervised scenario.

According to the numbers of source and target domains, we can divide UDA into the following four scenarios: (1) One source domain and one target domain (1S1T); (2) Multiple source domains and one target domain (mS1T); (3) One source domain and multiple target domains (1SmT); (4) Multiple source domains and multiple target domains (mSmT). In these scenarios, 1S1T has been studied extensively and mS1T has also been received relatively a fewer attention, while 1SmT and mSmT almost have no concerns. Since mSmT is essentially the combination of mS1T and 1SmT, this paper mainly focuses on 1SmT, where Fig.1 shows the difference of the first three scenarios.

\begin{figure}
\centering
  \includegraphics[width=0.8\linewidth]{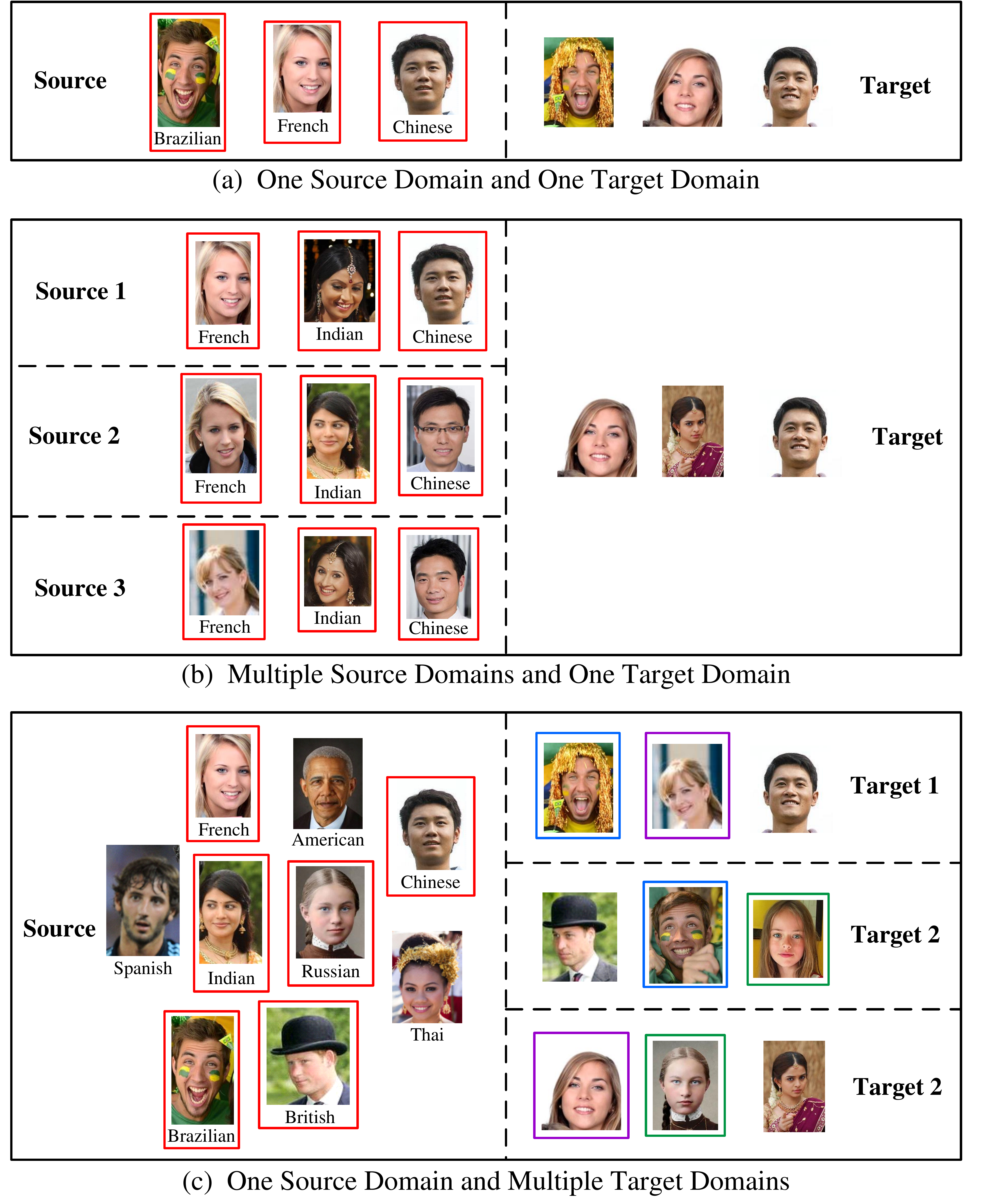}
  \caption{(a). UDA in 1S1T assumes that labeled source samples have different distributions with unlabeled target samples. (b). UDA in mS1T assumes that labeled source samples are collected from different source distributions, and they are different from target distributions. (c). UDA in 1SmT assumes that the distributions are different not only between the source domain and the individual target domains but also among the multiple target domains, moreover, the categories of source domain contain the categories of target domain, and these unlabeled target domains may not necessarily share the same categories.}
  \label{Fig:1}
\end{figure}

As mentioned above, until now, most of existing UDA works usually concern 1S1T, where the categories of source and target domains are assumed to be the same, and these approaches can be divided into two main types [2]. The one is \emph{instance-adaptation} approach [3], [4], [5] which is motivated by importance sampling, namely, the labeled samples in the source domain are re-weighted for the target domain to compensate for the distribution drift; the other is \emph{feature-adaptation} approach, [6], [7], [8], [9] which tries to discover a "good" feature representation by reducing the difference between the source and target domains [10]. Furthermore, state of the art feature-adaptation can further be subdivided into two main lines, namely instance alignment [11], [12], [13], [14] and distribution alignment [15], [16], [17], [18]. The former methods aim to seek a transformation matrix which aligns the source domain with the target one, while the latter ones aim to find a common latent subspace across the involved domains via various distance measures. Typically as maximum mean discrepancy (MMD) [19] which is often used to align distributions involved due to its efficiency and theoretical guarantee [15], [16].

In practice, since there may exist such cases that the categories of single source domain incompletely cover the ones of target domain, or completely cover the ones of target domain but the source samples are too scarce to transfer sufficient knowledge for the target domain. As a result, mS1T is adopted to solve such problems by using the complementary knowledge in multiple source domains with different distributions. However, the distribution discrepancy of these multiple source domains may lead to inferior result if just naively combining all source domain samples as a whole [20]. Subsequently, a method [21] was developed by the linearly weighted combination of the source distributions and has been proved better than the naive solution. The method [20] can be applied to the following two existed mS1T scenarios. The first scenario is a normal setting where multiple source domains and single target domain share the same categories, for which the instance-adaptation [22], [23] and the feature-adaptation [24] approaches can be adopted to solve the domain shift. The second one appears as a new scenario where the categories of individual source domains do not cover the target categories while these multiple source domains may not completely share their categories (called category shift) [25], for which the feature-adaptation [26] and the deep network [25] approaches can be used to battle the double shifts of domain and category among multiple source domains.

Besides UDA in 1S1T and mS1T, 1SmT likewise appears in practical application. One example is racial recognition as shown in Fig.1(c), where the race of everyone in the source domain is known but the race of three target domains is uncharted. Our goal is to recognize three related tasks, in which these data are sampled from different distributions. Obviously, there exist some implicit connections among three target domains, which is different from the previously mentioned new mS1T scenario. By contrast, such 1SmT has multiple related target domains to be learned, where we do not know whether their categories are overlapped or shared because of lacking labels. Thus, 1SmT is harder to find the connections among multiple target domains, leading to a more challenge than mS1T.

Through the analysis above, we can find that 1SmT transfers knowledge not only from the source domain to the individual target domains but also among target domains, where the lack of supervised information in these target domains poses a great difficulty in seeking inherent connections and transferable knowledge. Such a knowledge transfer among unsupervised target domains is called \emph{whole unsupervised domain adaptation} (WUDA), and to the best of our knowledge, until now, there is just one work [27] proposed which the authors develop transfer fuzzy c-means clustering (TFCM) and transfer fuzzy subspace clustering (TFSC). However, 1SmT needs to settle not only the UDA but also the WUDA, where the learning task under study is more complicated. Therefore, we propose a model parameter adaptation framework (PA-1SmT) for this scenario to transfer knowledge through adaptive learning of a common model parameter dictionary, and in turn, use the common model parameter dictionary to sparsely represent individual target model parameters. Fig.2 shows the framework of PA-1SmT.

In summary, the main contributions of this paper are as follows:

\hangafter 1 
\hangindent 2.6em 
$\bullet$ We present a new UDA scenario called 1SmT, which not only can solve 1S1T where the source categories are more than the target ones but also complements the existing UDA scenarios, i.e., 1S1T and mS1T, and provides basic for extending to mSmT.

\hangafter 1
\hangindent 2.6em
$\bullet$ We propose a model parameter adaptation framework to solve such a new scenario, which not only encodes a set of model parameter dictionary between the source domain and the individual target domains but also constructs a target-common model parameter dictionary among the multiple target domains. Then these dictionary coefficient matrixes are constrained by introducing \emph{L}$_{2,1}$ Norm so that the individual target model parameters can be sparsely represented by target-common model parameter dictionary, which attains model parameter adaptation among these domains.

\hangafter 1
\hangindent 2.6em
$\bullet$ Our approach differs from the existing UDA approaches completely in the knowledge transfer, such as instance adaptation and feature adaptation. These two approaches need to know the source samples when transferring knowledge from the source domain to the target domain. By contrast, our approach transfers knowledge via model parameter rather than data itself, thus can directly use for UDA about privacy protection.

\hangafter 1
\hangindent 2.6em
$\bullet$ We conduct experiments on three well-known domain adaptation benchmarks, and validate our approach in 1S1T and 1SmT scenarios. The empirical results demonstrate that the proposed approach achieves the state-of-the-art clustering performance than other methods.

\begin{figure}
\centering
  \includegraphics[width=0.9\linewidth]{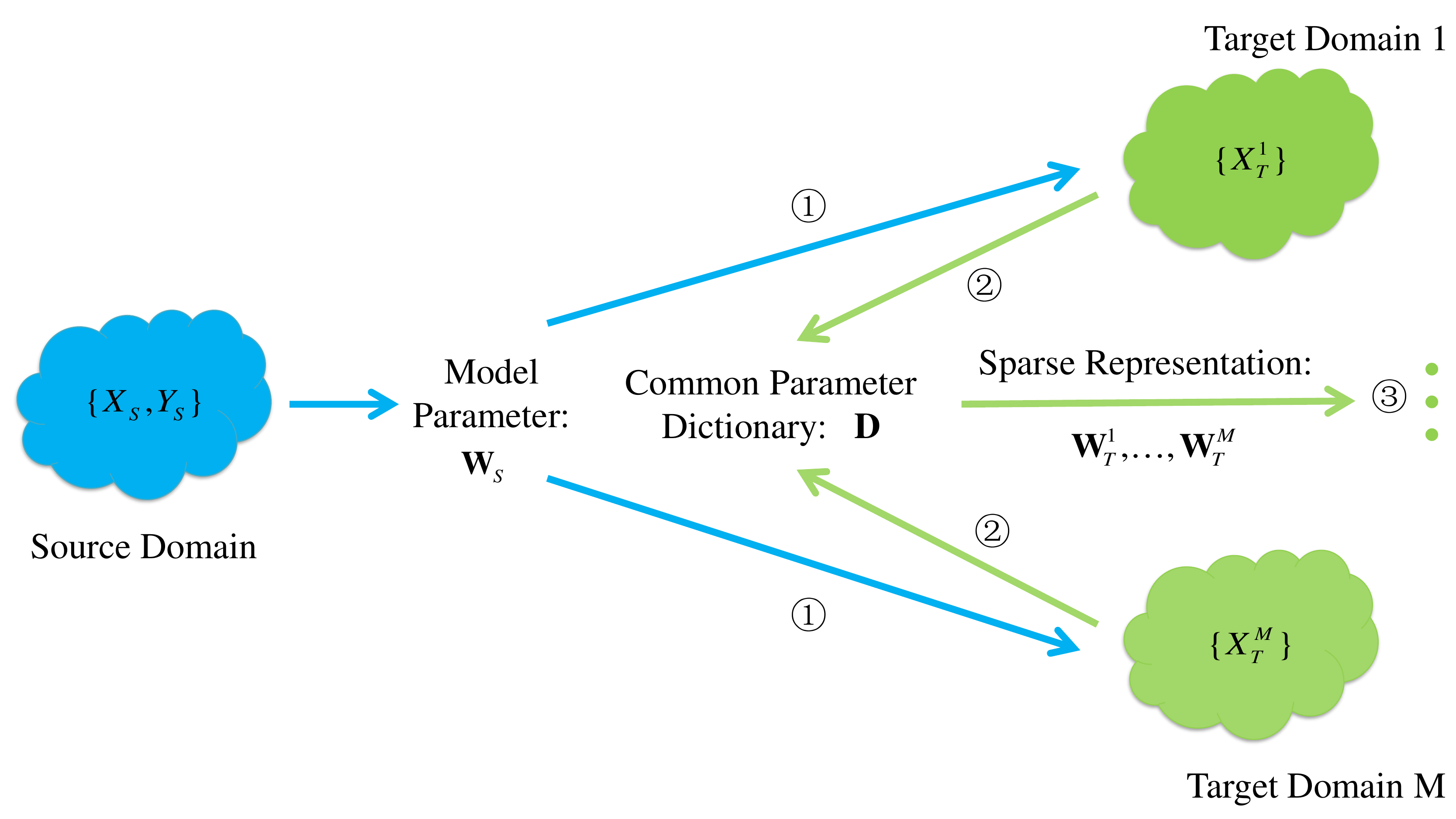}
  \caption{The framework of PA-1SmT. \textcircled{1} represents the model parameter transfer from the source domain to the target domain, \textcircled{2} represents the learning of a common model parameter dictionary, \textcircled{3} represents getting the target model parameters by sparse representation. $\mathbf{W}$ is model parameter of classifier/clusterer $f(x)=\mathbf{W}^Tx$, $\mathbf{D}$ is target-common model parameter dictionary among multiple target domains.}
  \label{Fig:2}
\end{figure}

The rest of this paper is organized as follows. We mention a number of related works in Section II. The proposed framework is detailed in Section III. Section IV describes the optimization procedure used in the proposed approach and analyzes the complexity and convergence. And section V highlights the experimental evaluations which is followed by concluding remarks and discussions on future endeavors that can be carried out based on the proposed framework in Section VI.

\section{Related Work}
In this section, we discuss three related topics, including unsupervised domain adaptation, whole unsupervised domain adaptation and soft large margin clustering.

\textbf{Unsupervised Domain Adaptation.} Recently, a great large of efforts have been made for UDA respectively based on non-deep architecture and deep architecture [25], [26], [28], [29], [30], [31], [32], [33], [34], [35], [36], which mainly contain the instance and the feature adaptations. For the instance adaptation method, a typical example is Kullback-Leibler Importance Estimation Procedure (KLIE) [4], which tries to reweight the source data and then use for the target one. Another method, i.e., feature adaptation, devotes to mitigating the domain shift by distribution alignment, such as MMD [15], [16], Correlation Alignment (CORAL) [17], deep CORAL [28], [29] and Central Moment Discrepancy (CMD) [18]. By contrast, our approach transfers the knowledge by adapting the model parameter instead of aligning the domain distributions. In this process, the improper model parameters will be automatically removed by an optimization. On the other hand, except for these methods above, recently, UDAs based on adversarial network [25], [34], [35], [36] use adversarial technology to find transferable features suitable for two domains. In this paper, our approach adopts the non-deep architecture to demonstrate conceptually our proposal, however, such a methodology can likewise be adapted to deep architectures by e.g., using the weights of the last layer, or the weights of corresponding layers in different domains as the model parameters to learn model parameter dictionaries for knowledge transfer.

\textbf{Whole Unsupervised Domain Adaptation.} The UDA mentioned above mainly considers the labeled source domain and the unlabeled target domain. When the source domain also has no labels, such a scenario becomes WUDA, which is similar to another sub-branch of transfer learning, i.e., unsupervised transfer learning (UTL). However, according to the definition in [10], WUDA is different from UTL in problem solving. Specifically, WUDA concerns on different domains whereas UTL puts more attention on different tasks. On the other hand, since the lack of supervised information in two domains makes it hard to find the transferable knowledge. To date, there are still relatively a few researches for such two scenarios. Among these works, TFCM and TFSC [27] belong to WUDA, which directly transfer the knowledge from the cluster centers obtained in the source domain to another clustering task. While self-taught clustering (STC) [37] and transfer spectral clustering (TSC) [38] belong to UTL, both adopt the common feature representation to transfer knowledge.

\textbf{Soft Larget Margin Clustering.} In this paper, we use soft large margin clustering (SLMC) [39] as the base function, because SLMC clusters in the label space by adopting the classification learning principle, which fits well our framework based on the model parameter dictionary learning. Moreover, SLMC can attain a better clustering performance than other traditional clustering algorithms, such as K-means [40], fuzzy \emph {c}-means (FCM) [41] and maximum margin clustering (MMC) [42] etc.. In detail, SLMC uses the encoding for the predefined \emph {C} cluster labels, and acquires result by the decision function and the soft membership. Specifically, given a set of data $ \textbf X=\{x_i\}_{i=1}^n $ where $ x_i\in \mathbb{R}^d $. Let $ f(x)=\textbf W^Tx$ denotes the decision function for a \emph {C}-cluster clustering, where $ \textbf W\in \mathbb{R}^{d \times C}$ is a weight parameter matrix. The optimization problem of SLMC can be formulated as:
\begin{eqnarray}
&&\mathop{\min}_{u_{ki}}\mathop{\min}_{\mathbf W} {1 \over 2}\| \mathbf W\|_F^2+{\lambda \over 2}\sum _{k=1}^\emph {C}\sum _{i=1}^\emph {n}u_{ki}^2\| \mathbf W^Tx_i-l_k\|_2^2 \nonumber \\
&&{s.t.}\sum _{k=1}^\emph {C}u_{ki}=1 \nonumber \\
&&0 \leq u_{ki} \leq 1,\; \forall k=1...\emph {C},i=1...\emph {n}
\end{eqnarray}
where $\mathbf U=[u_{ki}]_{\emph {C} \times n}$ denotes the soft partition matrix, in which $u_{ki}\in [0,1]$ represents the soft membership of $x_i$ to the \emph {k}th cluster. $\{l_1...l_\emph {C}\}$ denote the given encodings for the \emph {C} cluster respectively, and each $l_k=[0,...,0,1,0,...,0]^T \in \mathbb{R}^\emph {C}$ corresponding to the \emph {k}-th cluster , where the \emph {k}-th entry of $l_k$ is set to 1 and the rest to 0.

\section{Model Parameter Adaptation Framework for Unsupervised Domain Adaptationk}
In this section, we formulate the proposed model parameter adaptation framework (PA-1SmT), and model the PA-1SmT based on model parameter dictionary learning. In the following, we are going to explain this problem in two aspects and finally give the unified object function.

\subsection{Problem Setting}
\textbf{Normal UDA.} In the context of traditional UDA, we usually assume that the distributions are different between the source and the target domains($P_S(x,y)$ $\not= P_T(x,y)$). On the other hand, the categories of the source and target domains are same, namely, ${\cal C}_S ={\cal C}_T$.

\textbf{1SmT UDA.} Under the 1SmT scenario, the labeled source domain samples $\{(x_{Si},y_{Si})\}_{i=1}^{n_s}$  are drawn from $P_S(x,y)$, where $x_{Si}\in \mathbb{R}^d$ represents \emph {i}-th sample in the source domain and $y_{Si}$ is the corresponding label. Besides, we have a set of unlabeled data $\{\mathbf X_T^j\}_{j=1}^M$ in M different target domains, where $\mathbf X_T^j=\{x_{Ti}^j\}_{i=1}^{n_t^j}$ represents \emph {j}-th target domain samples and $x_{Ti}^j \in \mathbb{R}^d$, and the distribution of every target domain is $P_T^j(x,y)$. Thus, the assumption of distribution in this paper is as follows: the distributions are different not only between the source domain and the individual target domains but also among multiple target domains: $P_S(x,y)\not=P_T^j(x,y)$ and $P_T^j(x,y)\not= P_T^k(x,y)$ ($j\not=k$). In addition, contrasting to the normal UDA, our scenario is different in category sharing, i.e., the source categories contain all target categories, and these unlabeled target domains may not necessarily share same categories. In details, given a source category set ${\cal C}_S$ and M target category sets $\{{\cal C}_T^j\}_{j=1}^M$, their corresponding category numbers are $C_S$ and $\{C_T^j\}_{j=1}^M$ respectively, where ${\cal C}_T^j$ represents  \emph {j}-th target category. Let the union of all categories in multiple target domains are subset of the source categories ($\bigcup_{j=1}^M {\cal C}_T^j \subseteq {\cal C}_S$ and $\bigcap_{j=1}^M {\cal C}_T^j \not =\bigcup_{j=1}^M {\cal C}_T^j$). There may exist the situation in the target domains:
$ \bigcap_{j=1}^M {\cal C}_T^j ={\cal C}_c $ or $\bigcap_{j=1}^M {\cal C}_T^j =0 \land {\cal C}_T^j \cap {\cal C}_T^k={\cal C}_{c0}$, where ${\cal C}_c$ and ${\cal C}_{c0}$ represent common category in all or partial target domains respectively.

\subsection{Model Parameter Dictionary Learning}
\subsubsection{Learning from Source Domain to Target Domain}
At the beginning, we get the model parameter $\mathbf W_S \in \mathbb{R}^{d \times C_S}$ from the source domain, and learn the model parameter dictionary between the source domain and the individual target domains. In this process, we regard the target model parameter $\mathbf W_T^j \in \mathbb{R}^{d \times C_T^j}$ as dictionary of source model parameter because the source categories contain the target ones. Then, for \emph {j}-th target domain, the objective function of optimizing the model parameters is formulated as follows:
\begin{eqnarray}
&&\mathop{\min}_{\mathbf W_T^j,\mathbf V_j}\|\mathbf W_S-\mathbf W_T^j \mathbf V^j\|_F^2 + \eta \|\mathbf V^j\|_{2,1}
\end{eqnarray}
where $\mathbf V^j=[v_1^j,...,v_{C_S}^j]\in \mathbb{R}^{C_T^j \times C_S}$ is sparse representation of $\mathbf W_S$, $\eta$ is the tradeoff hyperparameter. And for $\mathbf M \in \mathbb{R}^{n \times m}$, we define $\|\mathbf M\|_{2,1}=\sum_{i=1}^n \|\mathbf m^i\|_2$, $\|\mathbf m^i\|$ is the \emph{i}-th row of the $\mathbf M$. Thus $\mathbf W_S$ can be sparsely represented by $\mathbf W_T^j$.

In practice, the first term in Eq.(2) is the model parameter dictionary learning from a source domain to the individual target domains, which is equal to learning a linear transformation to align the source model parameter to the target model parameter instead of aligning the existing often used subspaces and distributions. The second term is the sparsity constraint of dictionary coefficient matrix. We called these two terms as \textbf{S-T adaptive} terms, that is because they attain knowledge transfer by these model parameters from the source domain to the target domains.

\subsubsection{Learning among Multiple target Domains}
After the model parameter dictionary learning between the source and target domains, we then learn a target-common model parameter dictionary among multiple target domains, which falls in WUDA scenario. As mentioned before, it is more challenging than UDA for which we still follow the same modeling line as in the subsection 3.2.1 to formulate the objective to learn
\begin{eqnarray}
&&\mathop{\min}_{\mathbf D,\mathbf V_T^j}\|\mathbf W_T^j-\mathbf D \mathbf V_T^j\|_F^2 + \eta \|\mathbf V_T^j\|_{2,1}
\end{eqnarray}
where $\mathbf D =[d_1,...,d_r]\in \mathbb{R}^{r\times C_T^j}$ is a common dictionary of multiple target model parameters, $\mathbf V_T^j$ is the sparse coding of the \emph{j}-th target model parameter. $\eta$ is the tradeoff hyperparameter. From Eq.(3), every target model parameter can be sparsely represented by $\mathbf D$, which attains so-desired model parameter adaptation and transfers the knowledge among these target domains.

Likewise, the first term in Eq.(3) tries to learn a common dictionary among multiple target model parameters, without any supervised information of target domains, which realizes the inherent  knowledge connections among multiple unlabeled target domains. The second term is the sparsity constraint, by which this common model parameter dictionary can sparsely represent the individual target model parameters. Thus these two terms are called as \textbf{T-T adaptive} terms. Here it is necessary to mention that our such an approach is new its own right and can have interest of independence for WUDA.

\subsubsection{Unified Objective Function}
Due to no labels in the target domains, we adopt SLMC as the base function to train these target domains. For the \emph {j}-th target domain, we try to simultaneously get the knowledge from the labeled source domain and other unlabeled target domains. Thus, we integrate all the ingredients represented in the previous subsections, \emph {i.e.}, Eq.(1), Eq.(2) and Eq.(3), then obtain our final unified PA-1SmT framework formulated as Eq.(4).
\begin{eqnarray}
&&\mathop{\min}_{\mathbf W_T^j,u_{ki}^j,\mathbf D, \mathbf V^j,\mathbf V_T^j} \sum_{j=1}^M \bigg \{{1 \over 2}\| \mathbf W_T^j\|_F^2 \nonumber \\
&&\ \ \ \ \ \ \ \ \ \ \ \ \ \ \ \ \ \ \ \ +{\lambda \over 2}\sum_{k=1}^{C_T^j}\sum_{i=1}^ {n_t^j}(u_{ki}^j)^2\| (\mathbf W_T^j)^Tx_i^j-l_k^j\|_2^2 \nonumber \\
&&\ \ \ \ \ \ \ \ \ \ \ \ \ \ \ \ \ \ \ \ +\ {\beta \over 2} \|\mathbf W_S-\mathbf W_T^j\mathbf V^j\|_F^2+{\gamma \over 2}\|\mathbf W_T^j-\mathbf D \mathbf V_T^j\|_F^2  \nonumber\\
&&\ \ \ \ \ \ \ \ \ \ \ \ \ \ \ \ \ \ \ \ +\ \eta \bigg (\|\mathbf V^j\|_{2,1}+\|\mathbf V_T^j\|_{2,1}\bigg )\bigg\}\nonumber \\
&&{s.t.}\sum _{k=1}^{C_T^j}u_{ki}^j=1 \nonumber \\
&&0 \leq u_{ki}^j \leq 1,\; \forall k=1...{C_T^j},i=1...\emph {n},j=1...\emph {M}
\end{eqnarray}
where $\beta$ and $\gamma$ are tradeoff hyperparameters, which respectively represent the transfer degrees from the source domain to the individual target domains and among multiple target domains.

Last but importantly, we need to state that although our framework is based on SLMC, for other methods, such as FCM with model parameters $\mathbf V$(a set of clustering centers) and its variants, MMC with model parameters $\mathbf W$, such a framework can similarly be applied. Therefore, the proposed PA-1SmT framework has a wider applicability.

\section{Optimization and Analysis}
\subsection{Optimization Algorithm}
The optimization function in Eq.(4) is not-convex w.r.t joint $(\mathbf W_T^j,u_{ki}^j,\mathbf D, \mathbf V_j, \mathbf V_T^j)$ but block convex [43], i.e., the objective function is convex in single block of variable. Fortunately, for such block-convex problem, the existing work [43] has allowed us to use an alternating iterative strategy to solve. Specifically, for the \emph{j}-th target domain, the optimization procedure can be decomposed into the following three parts: (1) updating $u_{ki}^j$ with fixed $\mathbf W_T^j,\mathbf D, \mathbf V_j, \mathbf V_T^j$; (2) updating $\mathbf W_T^j$ and $\mathbf D$ with fixed $u_{ki}^j,\mathbf V_j,\mathbf V_T^j$; (3) updating $\mathbf V_j$ and $\mathbf V_T^j$ with fixed $\mathbf W_T^j,u_{ki}^j,\mathbf D$. The details are as follows:

\textbf {(i) Solving $u_{ki}^j$:}
Given fixed $\mathbf W_T^j,\mathbf D, \mathbf V_j, \mathbf V_T^j$, Eq.(4) is reduced to:
\begin{eqnarray}
&&\mathop{\min}_{u_{ki}^j}\sum_{k=1}^{C_T^j}\sum_{i=1}^ {n_t^j}(u_{ki}^j)^2\| (\mathbf W_T^j)^Tx_i^j-l_k\|_2^2 \nonumber \\
&&{s.t.}\sum _{k=1}^{C_T^j}u_{ki}^j=1 \nonumber \\
&&0 \leq u_{ki}^j \leq 1,\; \forall k=1...{C_T^j},i=1...\emph {n},j=1...\emph {M}
\end{eqnarray}
Through adopting the Lagrange multiplier method, Eq.(5) can be redefined as:
\begin{eqnarray}
&&\mathcal{J}(u_{ki}^j)=\sum_{k=1}^{C_T^j}\sum_{i=1}^ {n_t^j}(u_{ki}^j)^2\| (\mathbf W_T^j)^Tx_i^j-l_k\|_2^2\nonumber \\
&&\ \ \ \ \ \ \ \ \ \ \ \ \ \ \ \ -\sum_{i=1}^{n_t^j} \xi_i^j(\sum_{k=1}^{C_T^j}u_{ki}^j-1)
\end{eqnarray}
The derivative of $\mathcal{J}$ w.r.t each $u_{ki}^j$ is
\begin{eqnarray}
\frac{\partial \mathcal{J}}{\partial u_{ki}^j}=2\| (f^j(x_i^j)-l_k\|_2^2\ u_{ki}^j- \xi_i^j
\end{eqnarray}
Here, combining the constrain $\sum_{k=1}^{C_T^j}u_{ki}^j=1$, and let $\frac{\partial \mathcal{J}}{\partial u_{ki}^j}=0$, we get the following closed-form solution:
\begin{eqnarray}
u_{ki}^j=\frac{\| f^j(x_i^j)-l_k\|_2^{-2}}{\sum_{r=1}^{C_T^j}\| (f^j(x_i^j)-l_r\|_2^{-2}}
\end{eqnarray}

\textbf {(ii) Solving $\mathbf W_T^j$ and $\mathbf D$:}
This part includes the optimization of the target model parameter and the target-common model parameter dictionary. Thus, we describe $\mathbf W_T^j$ and $\mathbf D$ as shown below:
\begin{eqnarray}
&&\mathop{\min}_{\mathbf W_T^j,\mathbf D} \sum_{j=1}^M \bigg \{\| \mathbf W_T^j\|_F^2+{\lambda}\sum_{k=1}^{C_T^j}\sum_{i=1}^ {n_t^j}(u_{ki}^j)^2\| (\mathbf W_T^j)^Tx_i^j-l_k\|_2^2 \nonumber \\
&&\ \ \ \ \ \ \ \ +\ {\beta} \|\mathbf W_S-\mathbf W_T^j\mathbf V^j\|_F^2+{\gamma}\|\mathbf W_T^j-\mathbf D \mathbf V_T^j\|_F^2\bigg\}\nonumber \\
\end{eqnarray}

(a). Fixing $\mathbf D$, Eq.(9) is translated to the optimization problem w.r.t $\mathbf W_T^j $
\begin{eqnarray}
&&\mathop{\min}_{\mathbf W_T^j}\mathcal {J}(\mathbf W_T^j)= \sum_{j=1}^M \bigg \{\| \mathbf W_T^j\|_F^2 \nonumber \\
&&\ \ \ \ \ \ \ \ +{\lambda}\sum_{k=1}^{C_T^j}\sum_{i=1}^ {n_t^j}(u_{ki}^j)^2\| (\mathbf W_T^j)^Tx_i^j-l_k\|_2^2 \nonumber \\
&&\ \ \ \ \ \ \ \ +\ {\beta} \|\mathbf W_S-\mathbf W_T^j\mathbf V^j\|_F^2+{\gamma}\|\mathbf W_T^j-\mathbf D \mathbf V_T^j\|_F^2\bigg\}\nonumber \\
\end{eqnarray}
Obviously, it is a quadratic convex problem, thus letting $\frac{\partial \mathcal{J}}{\partial \mathbf W_T^j}=0$, we have
\begin{eqnarray}
&&\bigg (\mathbf I+\lambda \sum_{k=1}^{C_T^j}\mathbf X^j \hat {\mathbf U}_k(\mathbf X^j)^T+\gamma \mathbf I\bigg )\mathbf W_T^j+\mathbf W_T^j \bigg (\beta \mathbf V^j(\mathbf V^j)^T \bigg )\nonumber \\
&&\ \ \ \ \ \ \  =\lambda \sum_{k=1}^{C_T^j}\mathbf X^j \hat {\mathbf U}_k (\mathbf L_k^j)^T+\beta \mathbf W_S(\mathbf V^j)^T+\gamma \mathbf D \mathbf V_T^j
\end{eqnarray}
Eq.(11) is a famous Sylvester equation [44]. Since we code the PA-1SmT based on python, the Sylvester equation $\mathbf {AX}+\mathbf {XB}=\mathbf Q$ can directly be solved by python function: $scipy.linalg.\_solvers.solve\_ sylvester(\mathbf A,\mathbf B,\mathbf Q)$.

In Eq.(11), each $\mathbf L_k^j$ is a $C_T^j \times n_t^j$ matrix with the \emph{k}-th row being an all-one vector, and the rests being all-zero vectors, $\mathbf U_k$ denotes the \emph{k}-th row of $\mathbf U$, and $\hat {\mathbf U}_k$ denotes a diagonal matrix with the diagonal elements equaling to the squared values of the entries in $\mathbf U_k$.

(b). Fixing $\mathbf W_T^j$, Eq.(9) is reduced to:
\begin{eqnarray}
&&\mathop{\min}_{\mathbf D}\mathcal {J}(\mathbf D)= \sum_{j=1}^M \|\mathbf W_T^j-\mathbf D \mathbf V_T^j\|_F^2\nonumber \\
\end{eqnarray}
Similarly, let $\frac{\partial \mathcal{J}}{\partial \mathbf D}=0$, then the closed-form solution of $\mathbf D$ is obtained as:
\begin{eqnarray}
\mathbf D=\sum_{j=1}^M\mathbf W_T^j(\mathbf V_T^j)^T\bigg (\sum_{j=1}^M \mathbf V_T^j(\mathbf V_T^j)^T\bigg)^{-1}
\end{eqnarray}

\textbf {(iii) Solving $\mathbf V_j$ and $\mathbf V_T^j$:}
Given $\mathbf W_T^j,u_{ki}^j,\mathbf D$, the objective function Eq.(4) can be rewritten as follows:
\begin{eqnarray}
&&\mathop{\min}_{\mathbf V^j,\mathbf V_T^j} \sum_{j=1}^M \bigg \{{\beta \over 2} \|\mathbf W_S-\mathbf W_T^j\mathbf V^j\|_F^2+{\gamma \over 2}\|\mathbf W_T^j-\mathbf D \mathbf V_T^j\|_F^2\nonumber\\
&&\ \ \ \ \ \ \ \ \ +\ \eta \bigg (\|\mathbf V^j\|_{2,1}+\|\mathbf V_T^j\|_{2,1}\bigg )\bigg\}\nonumber \\
\end{eqnarray}
Similarly, we fix $\mathbf V_j$ and $\mathbf V_T^j$ respectively to optimize other variables. Since they are of the same form, we only optimize $\mathbf V_j$ while omitting the other solving details to avoid redundance.
\begin{eqnarray}
&&\mathop{\min}_{\mathbf V^j}\mathcal {J}(\mathbf V^j)=\sum_{j=1}^M \bigg \{{\beta \over 2} \|\mathbf W_S-\mathbf W_T^j\mathbf V^j\|_F^2\nonumber\\
&&\ \ \ \ \ \ \ \ \ \ \ \ \ \ \ \ \ \ \ \ \ \ \ \ \ \ \ \ \ \ \ +\eta \|\mathbf V^j\|_{2,1}\bigg\}
\end{eqnarray}
To optimize Eq.(15) which involves the non-smooth term $\|\bullet\|_{2,1}$, thus we resort to the technique in [45], taking the derivative of $\mathcal {J}(\mathbf V^j)$ w.r.t $\mathbf V^j$, consequently we have
\begin{eqnarray}
&&\frac{\partial \mathcal{J}}{\partial \mathbf V^j}=-\beta (\mathbf W_T^j)^T(\mathbf W_S-\mathbf W_T^j\mathbf V^j)+2\eta\mathbf M^j\mathbf V^j
\end{eqnarray}
and set its derivative to zero, thus we have
\begin{eqnarray}
\mathbf V^j=\bigg (\beta (\mathbf W_T^j)^T\mathbf W_T^j+2\eta \mathbf M^j\bigg )^{-1}\beta (\mathbf W_T^j)^T\mathbf W_S
\end{eqnarray}
Similarly, the closed-form solution for $\mathbf V_T^j$ is
\begin{eqnarray}
\mathbf V_T^j=\bigg (\gamma \mathbf D^T\mathbf D+2\eta \mathbf M_T^j\bigg )^{-1}\gamma \mathbf D^T\mathbf W_T^j
\end{eqnarray}
where $\mathbf M^j$ and $\mathbf M_T^j$ are both diagonal matrixes with the \emph{i}-th diagonal elements respectively as:
\begin{eqnarray}
&&m_{ii}=\frac{1}{2\|(v^j)^i\|_2}\ ,\ (m_T)_{ii}=\frac{1}{2\|(v_T^j)^i\|_2}
\end{eqnarray}
where $(v^j)^i$ and $(v_T^j)^i$ denote the \emph{i}-th rows of $\mathbf V^j$ and $\mathbf V_T^j$ respectively.

\subsection{Complexity and Convergence Analysis}
\begin{algorithm}[h]
\caption{PA-1SmT Algorithm}
\label{alg1}
\begin{algorithmic}[1]
\REQUIRE

$\{\mathbf X_T^j\}_{j=1}^M$: M target domain datasets\\
$\ \ \ \ \ \ \mathbf W_S$: Model parameter of source domain \\
$\ \ \ \ \ \ \lambda , \beta ,\gamma, \eta$: Hyperparameters \\
$\ \ \ \ \ \ r$: Common dictionary size\\
$\ \ \ \ \ \ \{C_T^j\}_{j=1}^M$: Category numbers\\
\ENSURE

$\{\mathbf U^j\}_{j=1}^M$: Soft partition matrixes\\
\STATE Initialize $\{\mathbf U^j\}_{j=1}^M$;
\STATE Initialize $\mathbf D,\  \{\mathbf V_T^j\}_{j=1}^M,\  \{\mathbf V^j\}_{j=1}^M$;
\STATE Initialize $\{\mathbf M^j\}_{j=1}^M,\ \{\mathbf M_T^j\}_{j=1}^M$ as an identity matrix;
\STATE Set the initial objective function value to infinity, i.e., $J_0$=INF;
\REPEAT
\STATE Update $\{\mathbf W_T^j\}_{j=1}^M$ by Eq.(11);
\STATE Update $\{\mathbf U^j\}_{j=1}^M$ by Eq.(8);
\STATE Update $\{\mathbf V_T^j\}_{j=1}^M$ and $\{\mathbf V^j\}_{j=1}^M$ as follows:
\REPEAT
\STATE Calculate $\mathbf V^j$ and $\mathbf V_T^j$ by Eq.(17),(18);
\STATE Calculate diagonal matrix $ \mathbf M^j$ and $ \mathbf M_T^j$ by Eq.(19);
\UNTIL{Convergence};
\STATE Update $\mathbf D$ by Eq.(13);
\UNTIL{Convergence};
\STATE return $\{\mathbf U^1,...,\mathbf U^M\}$ and clustering results
\end{algorithmic}
\end{algorithm}

(1) \emph {Complexity Analysis}.
To analyze the time complexity of the proposed PA-1SmT algorithm, we first synthesize whole process for solving Eq.(4) in Algorithm 1. In each update of variables, their time complexities are mainly dominated by matrix multiplication and inverse operations. Thus, we provide some information before calculating the time complexity of PA-1SmT. The matrix multiplication costs $\mathcal {O}(nmk)$ for $m\times n$ and $n \times k$ matrixes, the matrix inverse costs $\mathcal {O}(r^3)$ for a r-order square matrix. Concretely, first, in Eq.(9), the cost of computing $\|f^j(x_i^j)-l_k\|_2^{-2}$ is $\mathcal {O}(dC_T^j)$. Thus, getting the membership matrix $\mathbf U^j$ costs $\mathcal {O}(n^jdC_T^{j\ 3})$ in the \emph {j}-th target domain. Second, the \emph {solve\_sylvester} function costs $\mathcal {O}(d^3)$ in updating Eq.(11), and the calculation of $\mathbf D$ costs $\mathcal {O}(\sum_{j=1}^MC_T^j(dr+r^2)+r^3+dr^2)$. Third, the time complexities in updating $\mathbf {V}^j$ and $\mathbf {V}_T^j$ are $\mathcal {O}(dC_T^jC_S+C_T^{j\ 3})$ and $\mathcal {O}(dr^2+drC_T^j+r^3)$ respectively. In general, $d\gg r,\ C_T^j$. Suppose $L$ is the iteration times of the outer loop, $n^{max}$ and $C_T^{max}$ are the largest sample and category numbers in the M target domains respectively, thus the total time complexity of PA-1SmT is $\mathcal {O}(Ld^3+Ln^{max}dC_T^{max\ 3})$.

(2) \emph{Convergence Analysis}. The convergence of the Algorithm 1 is summarized in the following theorem 1:

\textbf{Theorem 1:} The objective function value shown in Eq.(4) monotonically decreases until convergence by applying the proposed algorithm.

\textbf{Proof:} Suppose after the \emph{t}-th iteration, the objective function value is
$\mathcal {J}(u_{ki}^{j(t)}$, $\mathbf W_T^{j(t)}$,$\mathbf D^{(t)}$,$\mathbf V^{j(t)}$,$\mathbf V_T^{j(t)})$. Then in the next iteration, we fix $\mathbf W_T^{j(t)}$, $\mathbf D^{(t)}$, $\mathbf V^{j(t)}$, $\mathbf V_T^{j(t)}$, due to that objective function is convex in $u_{ki}^j$, thus we have
\begin{eqnarray}
&&\mathcal {J}(u_{ki}^{j(t+1)},\mathbf W_T^{j(t)},\mathbf D^{(t)},\mathbf V^{j(t)},\mathbf V_T^{j(t)})\leq \nonumber \\
&&\ \ \ \ \ \ \ \ \ \ \ \ \ \ \mathcal {J}(u_{ki}^{j(t)},\mathbf W_T^{j(t)},\mathbf D^{(t)},\mathbf V^{j(t)},\mathbf V_T^{j(t)})
\end{eqnarray}
In the same manner, we respectively solve the optimal solutions of $\mathbf W_T^j$ and $\mathbf D$, and the following inequalities hold:
\begin{eqnarray}
&&\mathcal {J}(u_{ki}^{j(t+1)},\mathbf W_T^{j(t+1)},\mathbf D^{(t)},\mathbf V^{j(t)},\mathbf V_T^{j(t)})\leq\nonumber \\
&&\ \ \ \ \ \ \ \ \ \ \ \ \ \mathcal {J}(u_{ki}^{j(t+1)},\mathbf W_T^{j(t)},\mathbf D^{(t)},\mathbf V^{j(t)},\mathbf V_T^{j(t)})\\
&&\mathcal {J}(u_{ki}^{j(t+1)},\mathbf W_T^{j(t+1)},\mathbf D^{(t+1)},\mathbf V^{j(t)},\mathbf V_T^{j(t)})\leq\nonumber \\
&&\ \ \ \ \ \ \ \ \ \ \ \ \mathcal {J}(u_{ki}^{j(t+1)},\mathbf W_T^{j(t+1)},\mathbf D^{(t)},\mathbf V^{j(t)},\mathbf V_T^{j(t)})
\end{eqnarray}
For the variables $\mathbf V^j$ and $\mathbf V_T^j$, based on the Lemma 1 [45], we have:
\begin{eqnarray}
&&\|v^{(t+1)}\|_2-\frac{\|v^{(t+1)}\|_2^2}{2\|v^{(t)}\|_2} \leq \|v^{(t)}\|_2-\frac{\|v^{(t)}\|_2^2}{2\|v^{(t)}\|_2}
\end{eqnarray}
and resorting to the Theorem 1 [45], we have:
\begin{eqnarray}
&&\|\mathbf V^{j(t+1)}\|_{2,1} \leq \|\mathbf V^{j(t)}\|_{2,1} \nonumber \\
&&\|\mathbf V_T^{j(t+1)}\|_{2,1} \leq \|\mathbf V_T^{j(t)}\|_{2,1}
\end{eqnarray}
Similarly, the following inequalities hold:
\begin{eqnarray}
&&\mathcal {J}(u_{ki}^{j(t+1)},\mathbf W_T^{j(t+1)},\mathbf D^{(t+1)},\mathbf V^{j(t+1)},\mathbf V_T^{j(t)})\leq\nonumber \\
&&\ \ \ \ \ \ \mathcal {J}(u_{ki}^{j(t+1)},\mathbf W_T^{j(t+1)},\mathbf D^{(t+1)},\mathbf V^{j(t)},\mathbf V_T^{j(t)})\\
&&\mathcal {J}(u_{ki}^{j(t+1)},\mathbf W_T^{j(t+1)},\mathbf D^{(t+1)},\mathbf V^{j(t+1)},\mathbf V_T^{j(t+1)})\leq\nonumber \\
&&\ \ \ \ \ \ \mathcal {J}(u_{ki}^{j(t+1)},\mathbf W_T^{j(t+1)},\mathbf D^{(t+1)},\mathbf V^{j(t+1)},\mathbf V_T^{j(t)})
\end{eqnarray}
From Eqs. (20-22, 25, 26), we can see that the objective function value decreases after each iteration. Further, since the objective function itself is non-negative, thus it exists a lower bound. Therefore, the Algorithm 1 will be convergent, Theorem 1 has been proved.

\section{Experiments and Results}
In this section, we evaluate the performance of PA-1SmT in three different cross-domain datasets. The hyperparameter sensitivity is then evaluated and discussed. Afterwards, compared with some baselines, we report and discuss the results of clustering.
\subsection{Date Sets and Settings}
\textbf {Data Sets:}
In this paper, we select three classical datasets to evaluate our method. In these datasets, the distributions are different among their sub-datasets, i.e., $P_{sub-i}(x,y)\not=P_{sub-j}(x,y)$, which satisfies our problem settings. Then the datasets detail as follow:

(1) \emph{Office}+\emph{Caltech datasets}
\footnote{https://github.com/jindongwang/transferlearning/blob/master/\\
data/dataset.md\#office+caltech}: Office dataset is the visual domain adaptation benchmark data, which includes 10 common object categories in three different sub-datasets, i.e., Amazon, DSLR, and Webcam. Besides, Caltech is a standard dataset for object recognition. These four sub-datasets then are considered as four different domains, as shown in Fig.3. In this experiment, we adopt the public Office + Caltech datasets released by [9], where each domain contains 10 objective categories. These sub-datasets are based on SURF features have been made by [9], which are extracted and quantized into an 800-bin histogram with codebooks computed with K-means on a subset of images from Amazon. Then the histograms are standardized by z-score.
\begin{figure}[ht]
\begin{center}
  \includegraphics[width=0.8\linewidth]{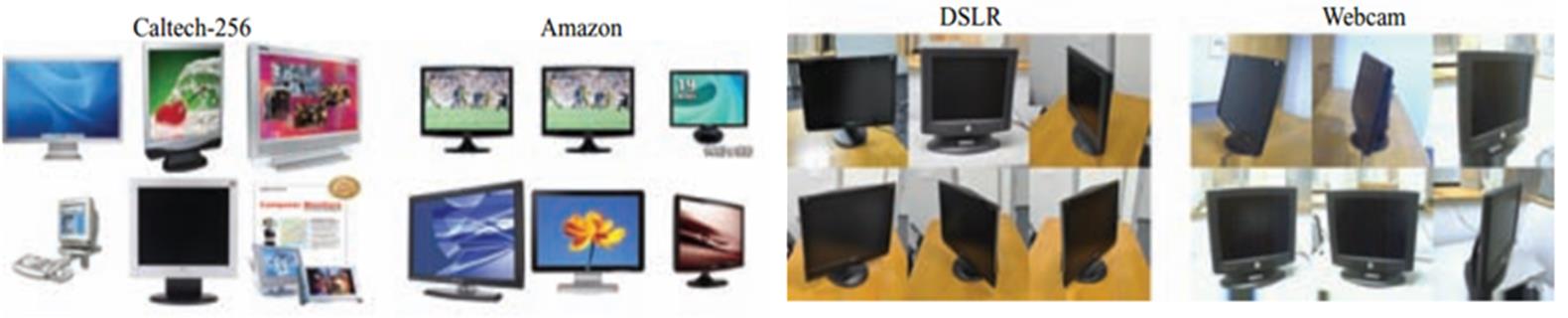}
  \caption{Images from domain adaptation benchmark data sets Office and Caltech}
  \label{Fig:3}
\end{center}
\end{figure}

(2) \emph{PIE dataset}
\footnote{https://github.com/jindongwang/transferlearning/blob/master/\\
data/dataset.md\#pie}: It contains 11554 face images with the resolution of 32 $\times$ 32 pixels from 68 individuals. These images have pose, illumination and expression changes. Fig.4 shows the pose variation of the PIE dataset [46]. In this experiment, five subsets of PIE (each corresponding to a distinct pose) are used to test different tasks. Specifically, five sub-datasets, i.e., PIE1 (C05, left pose), PIE2 (C07, upward pose), PIE3 (C09, downward pose), PIE4 (C27, front pose), PIE5 (C29, right pose), are constructed and the face images in each subset are taken under different illumination and expression conditions. And these sub-datasets are also based on SURF features.

\begin{figure}[ht]
\begin{center}
  \includegraphics[width=0.8\linewidth]{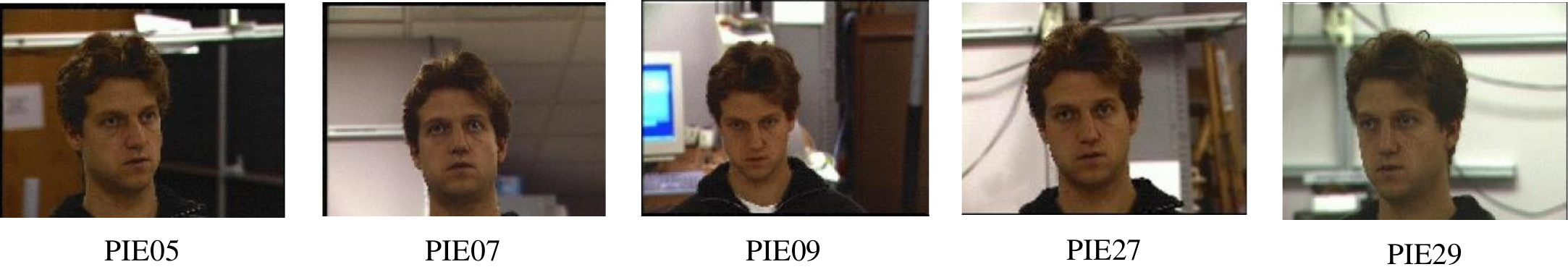}
  \caption{Images from domain adaptation benchmark data sets Office and Caltech}
  \label{Fig:4}
\end{center}
\end{figure}

(3) \emph{Extended Yale B dataset}
\footnote{http://www.cad.zju.edu.cn/home/dengcai/Data/FaceData.html}: It consists of 2,414 frontal face samples of 38 persons under various illumination conditions and each image has the resolution of $32\times 32$ pixels. Following [47], we divide the dataset into five sub-datasets(see as Fig.5). Sub-dataset 1 contains 266 images (seven images per subject) under normal lighting conditions. Sub-datasets 2 and 3, each consisting of 12 images per subject, characterize slight-to-moderate luminance variations, whereas sub-dataset 4 (14 images per person) and sub-dataset 5 (19 images per person) depicts severe light variations. We briefly name these sub-datasets as Y1, Y2, Y3, Y4, Y5 respectively.

\begin{figure}[ht]
\begin{center}
  \includegraphics[width=0.4\linewidth]{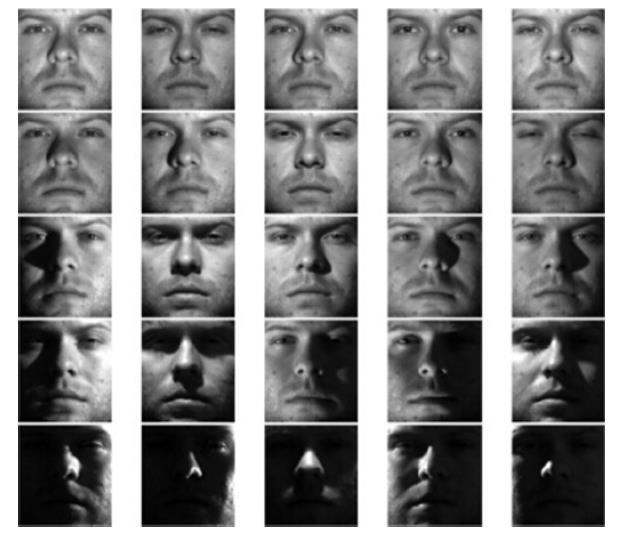}
  \caption{Starting from the top, each row shows images from subsets 1, 2, 3, 4, and 5, respectively.}
  \label{Fig:5}
\end{center}
\end{figure}

\textbf {Compared Methods:} Since our framework is based on clustering, and such a scenario 1SmT has not been researched, we compare our method with unsupervised transfer learning (STC, TSC) and whole unsupervised domain adaptation (TFSC), which are listed below:

\hangafter 1 
\hangindent 2.6em 
$\bullet$ \textbf{SLMC} [40]. SLMC is the base of our proposed approach. The comparison is to verify that our method could greatly avoid the negative transfer.

\hangafter 1 
\hangindent 2.6em 
$\bullet$ \textbf{STC} [38]. It aims to cluster a small collection of target unlabeled data with the help of a large amount of auxiliary unlabeled data. Then it tackles this problem based on co-clustering by clustering the target and auxiliary data simultaneously to allow the feature representation from the auxiliary data to influence the target data through a common set of features.

\hangafter 1 
\hangindent 2.6em 
$\bullet$ \textbf{TSC} [39]. It involves not only the data manifold information of the clustering task but also the feature manifold information shared between the related clustering tasks. Furthermore, it makes use of co-clustering to achieve and control the knowledge transfer between two tasks.

\hangafter 1 
\hangindent 2.6em 
$\bullet$ \textbf{TFSC} [28]. The idea of TFSC leverages knowledge in the source domain to help the limited and scared target data cluster, and then transfers knowledge through aligning the clustering centers of the source and the target domains.

The unsupervised transfer learning (STC, TSC) and whole unsupervised domain adaptation algorithms (TFSC) transfer knowledge from a unlabeled source domain, whereas our method is based on a labeled source domain and multiple unlabeled target domains. Therefore, for the fairness of the experiments, we add labels for the source domain in training. On the other hand, in order to eliminate the contingency of results, we run 10 times in every experiment, and then use the average value as the final results.

\textbf {Clustering Performance Measures:} In all experiments given below, we adopt the Normalized Mutual Information (NMI) [28] and Rand Index (RI) [28] to measure the clustering performance, where NMI measures the similar degree of the two clustering results, and RI computes the percentage of correct pair-wise relationships. For all these two measures, larger value means better cluster performance.

In the experiment, we use KPCA [47] to preprocess different features. In the KPCA preprocessing, we adopt the Gaussian kernel, i.e., $k(\mathbf x_i, \mathbf x_j)$=exp$(-\|\mathbf x_i-\mathbf x_j\|^2/(2{\omega}^2))$, where the hyperparameter $\omega$ is empirically set as the mean of the distances between all training sample pairs.

\subsection{Experimental Results}
1)\emph {Office+Caltech datasets:} For sake of the knowledge transfer in clustering, we use the Caltech and Amazon datasets with a relatively large sample size as the source domain respectively, and use the Webcam and Dslr as the target domains. Besides, in the scenario settings, the source domain categories contain the target ones, meanwhile, these unlabeled target domains may not necessarily share the same categories. Thus, we firstly deal with the original datasets before training, i.e., for the Webcam and Dslr datasets, we choose 6 categories and then share 4 categories between these two datasets. In addition, in the experiments, A$^1 \rightarrow$W, D and C$^1 \rightarrow$W, D represent the knowledge transfer in two target domains. Similarly, A$^1 \rightarrow$W, D , C$^2$ and C$^1 \rightarrow$W, D, A$^2$ are for three target domains.

\begin{table}[ht]
\begin{center}
\footnotesize
\caption{Average NMI of All Compared Methods on Office+Caltech Datasets }
\label{Table 1}
\setlength{\tabcolsep}{1.5mm}{
\begin{tabular}{|c|c|c|c|c|c|c|c|}
\hline
\multirow{2}{*}{\begin{tabular}[c]{@{}c@{}}\\$D_S$\end{tabular}} & \multirow{2}{*}{\begin{tabular}[c]{@{}c@{}}\\$D_T$\end{tabular}} & baseline & \multicolumn{4}{c|}{1S1T}                                                                          & 1SmT            \\ \cline{3-8}
                    &                     & SLMC     & TSC             & STC             & TFSC   & \begin{tabular}[c]{@{}c@{}}single\\ ours\end{tabular} & ours            \\ \hline
\multirow{2}{*}{A$^1$} & W                   & 0.1222   & 0.168           & 0.1759          & 0.1628 & \underline{\textit{0.26}}                                         & {\textbf{0.3011}} \\
                    & D                   & 0.1569   & 0.231           & 0.2292          & 0.1716 & \underline{\textit{0.2483}}                                       & {\textbf{0.2753}} \\ \hline
\multirow{2}{*}{C$^1$} & W                   & 0.1222   & 0.1237          & 0.1189          & 0.1413 & \underline{\emph{0.1677}}                                                & {\textbf{0.2637}} \\
                    & D                   & 0.1569   & 0.1852          & 0.1856          & 0.1849 & \underline{\textit{0.1986}}                                       & {\textbf{0.2897}} \\ \hline
\multirow{3}{*}{A$^1$} & W                   & 0.1222   & 0.168           & 0.1759          & 0.1628 & \underline{\textit{0.26}}                                         & {\textbf{0.3332}} \\
                    & D                   & 0.1569   & 0.231           & 0.2292          & 0.1716 & \underline{\textit{0.2483}}                                       & {\textbf{0.2866}} \\
                    & C$^2$                  & 0.0639   & 0.204           & \underline{\textit{0.2105}} & 0.196  & 0.1909                                                & {\textbf{0.2119}} \\ \hline
\multirow{3}{*}{C$^1$} & W                   & 0.1222   & 0.1237          & 0.1189          & 0.1413 & \underline{\textit{0.1677}}                                       & {\textbf{0.2706}} \\
                    & D                   & 0.1569   & 0.1852          & 0.1856          & 0.1849 & \underline{\textit{0.1986}}                                       & {\textbf{0.3009}} \\
                    & A$^2$                  & 0.0717   & {\textbf{0.2487}} & \underline{\textit{0.238}}  & 0.1683 & 0.2107                                                & 0.2353          \\ \hline
\end{tabular}}\\
\end{center}
\footnotesize{ The top NMI value is highlighted by bold font and the second best value is by italic with underline, which is the same with the following table except for Table 5. \\
The representation of number on the dataset, such as C$^1$, C$^2$, means that they have different categories, similarly hereinafter.}
\end{table}

\begin{table}[ht]
\begin{center}
\footnotesize
\caption{Average RI of All Compared Methods on Office+Caltech Datasets}
\label{Table 2}
\setlength{\tabcolsep}{1.5mm}{
\begin{tabular}{|c|c|c|c|c|c|c|c|}
\hline
\multirow{2}{*}{\begin{tabular}[c]{@{}c@{}}\\$D_S$\end{tabular}} & \multirow{2}{*}{\begin{tabular}[c]{@{}c@{}}\\$D_T$\end{tabular}} & baseline & \multicolumn{4}{c|}{1S1T}                                                                          & 1SmT            \\ \cline{3-8}
                    &                     & SLMC     & TSC             & STC             & TFSC   & \begin{tabular}[c]{@{}c@{}}single\\ ours\end{tabular} & ours            \\ \hline
\multirow{2}{*}{A$^1$} & W                   & 0.7146   & 0.7535          & 0.7542          & 0.7223 & \underline{\textit{0.7556}}                                       & {\textbf{0.7786}} \\
                    & D                   & 0.7213   & 0.7385          & \underline{\textit{0.7486}} & 0.7268 & 0.7464                                                & {\textbf{0.7499}} \\ \hline
\multirow{2}{*}{C$^1$} & W                   & 0.7146   & 0.7266          & 0.7263          & 0.7128 & \underline{\textit{0.7268}}                                       & {\textbf{0.7695}} \\
                    & D                   & 0.7213   & 0.7221          & 0.7215          & 0.7242 & \underline{\textit{0.7368}}                                       & {\textbf{0.7539}} \\ \hline
\multirow{3}{*}{A$^1$} & W                   & 0.7146   & 0.7535          & 0.7542          & 0.7223 & \underline{\textit{0.7556}}                                       & {\textbf{0.7816}} \\
                    & D                   & 0.7213   & 0.7385          & \underline{\textit{0.7486}} & 0.7268 & 0.7464                                                & {\textbf{0.7557}} \\
                    & C$^2$                  & 0.7002   & \underline{\textit{0.7443}} & {\textbf{0.7609}} & 0.7345 & 0.7396                                                & 0.7404          \\ \hline
\multirow{3}{*}{C$^1$} & W                   & 0.7146   & 0.7266          & 0.7263          & 0.7128 & \underline{\textit{0.7268}}                                       & {\textbf{0.7622}} \\
                    & D                   & 0.7213   & 0.7221          & 0.7215          & 0.7242 & \underline{\textit{0.7368}}                                       & {\textbf{0.7806}} \\
                    & A$^2$                  & 0.711    & \underline{\textit{0.7538}} & 0.7501          & 0.7161 & 0.749                                                 & {\textbf{0.7608}} \\ \hline
\end{tabular}}\\
\end{center}
\end{table}

\begin{figure}[ht]
\begin{center}
  \includegraphics[scale=0.3]{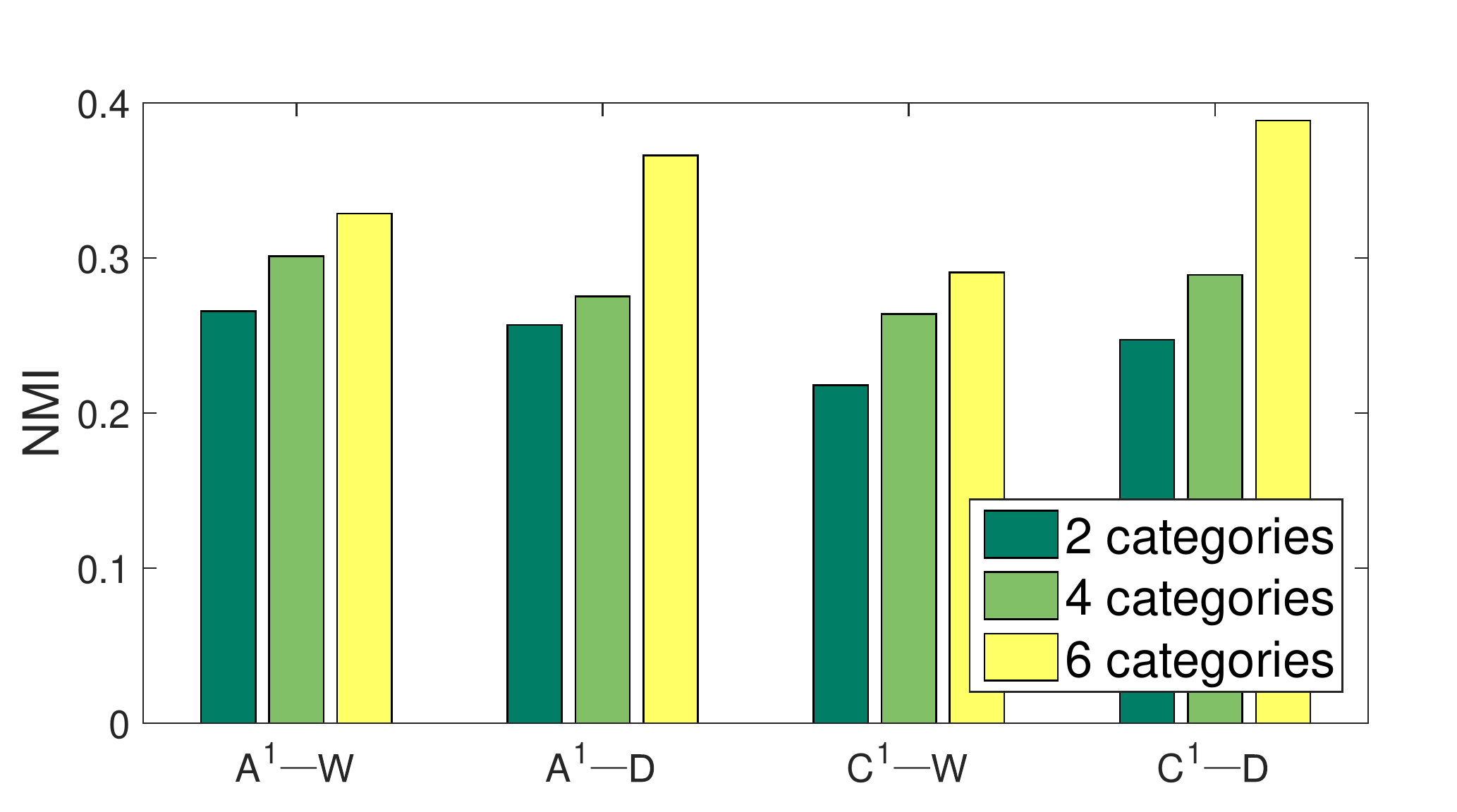}
  \caption{The comparison of different shared categories in two target domains. The dark green, light green and yellow columns respectively represent shared 2, 4 and 6 category numbers in the target domains.}
  \label{Fig:6}
\end{center}
\end{figure}

\begin{figure*}[ht]
  \centering
  \subfigure[]{
  \begin{minipage}{4cm}
  \centering
  \includegraphics[scale=0.22]{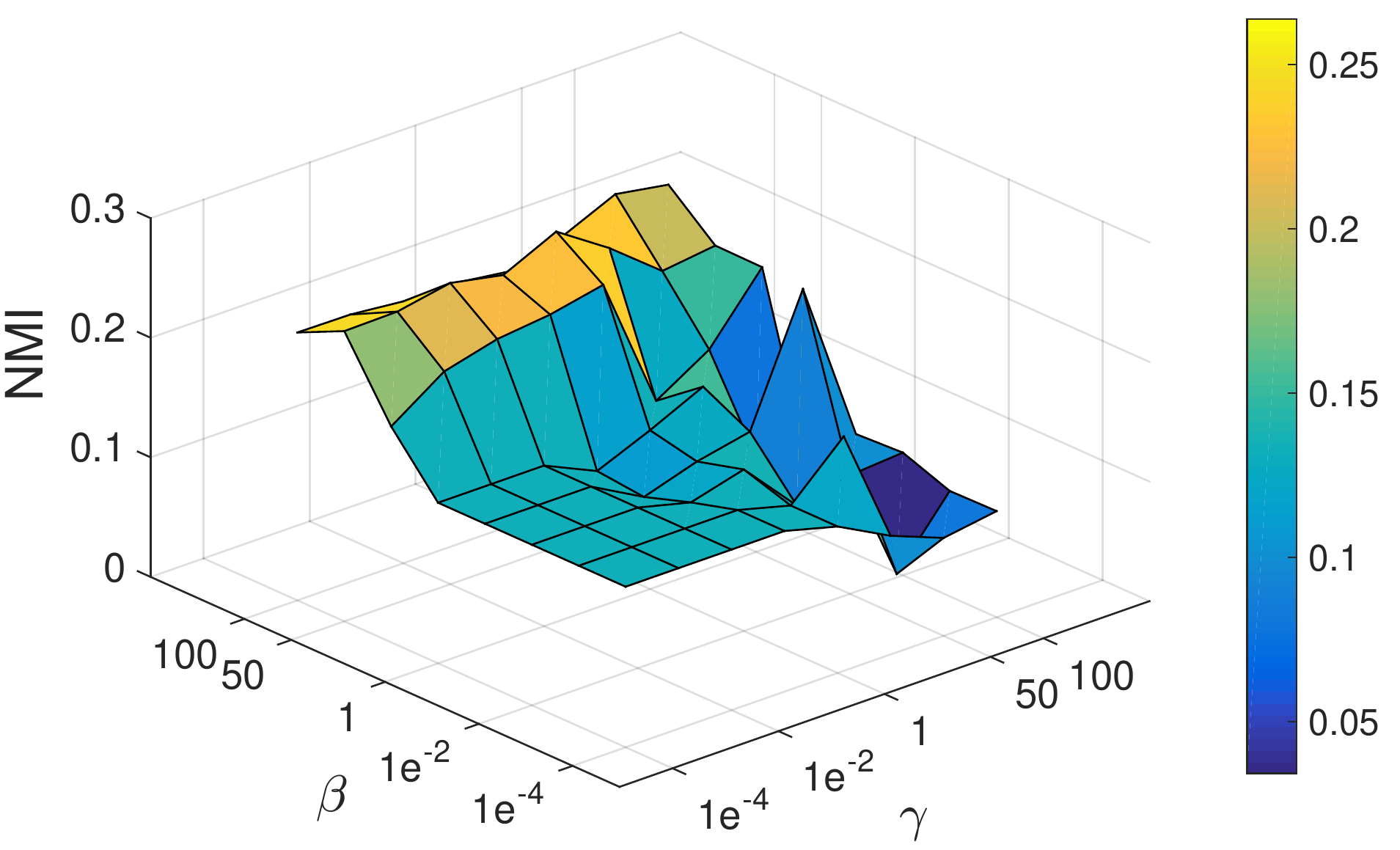}
  \end{minipage}
  }
  \subfigure[]{
  \begin{minipage}{4cm}
  \centering
  \includegraphics[scale=0.22]{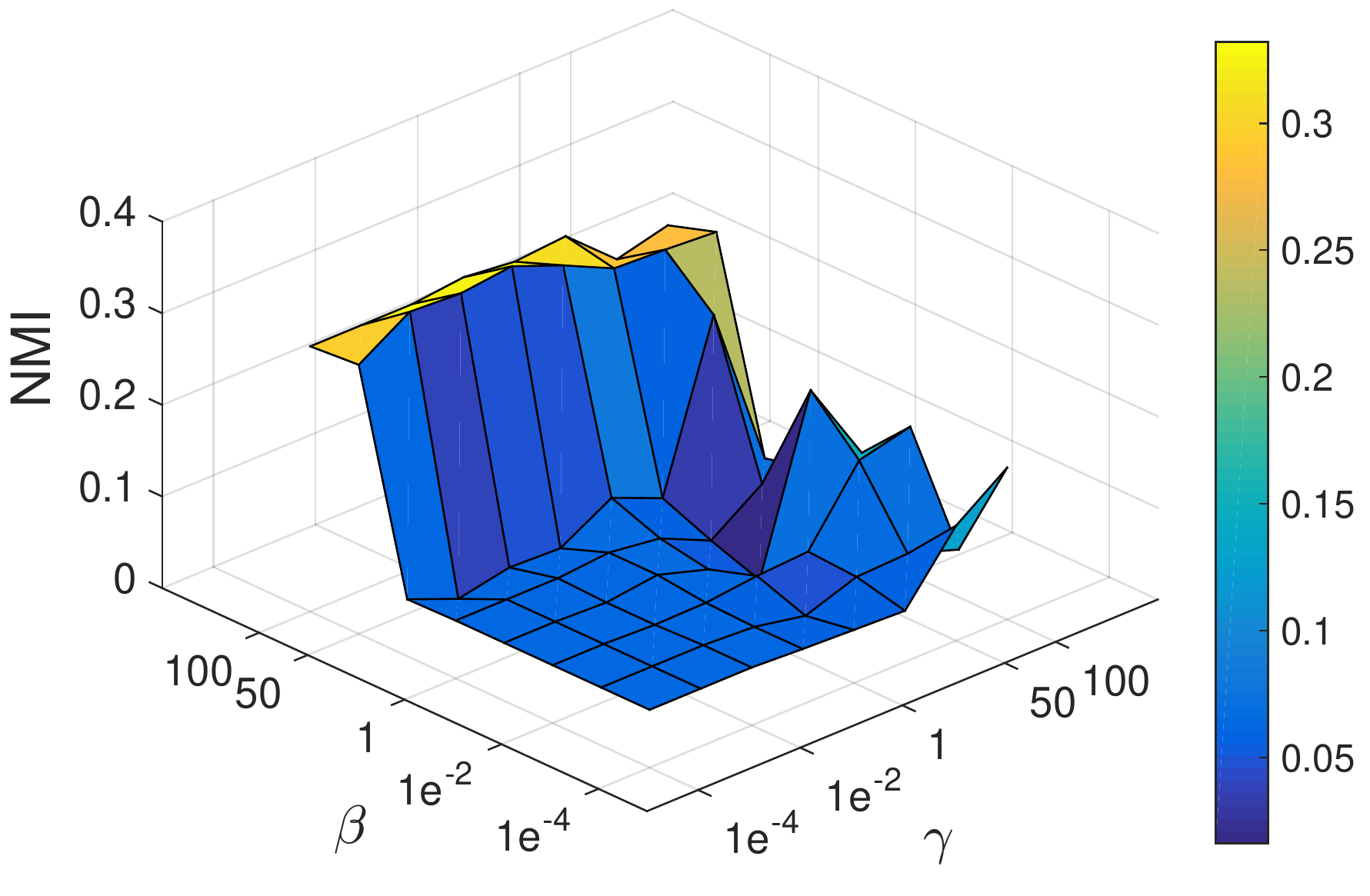}
  \end{minipage}
  }
  \subfigure[]{
  \begin{minipage}{4cm}
  \centering
  \includegraphics[scale=0.22]{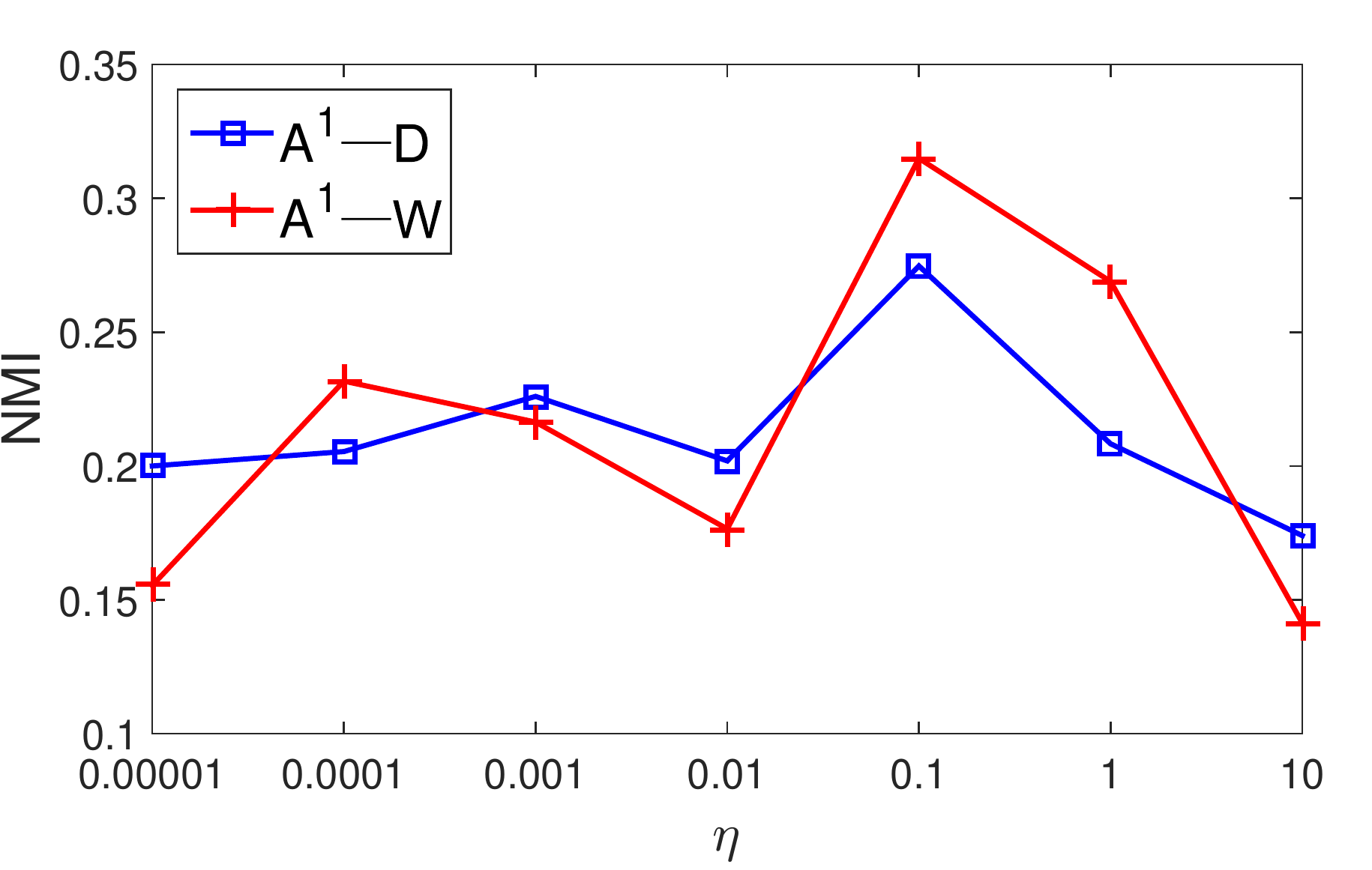}
  \end{minipage}
  }
  \subfigure[]{
  \begin{minipage}{4cm}
  \centering
  \includegraphics[scale=0.22]{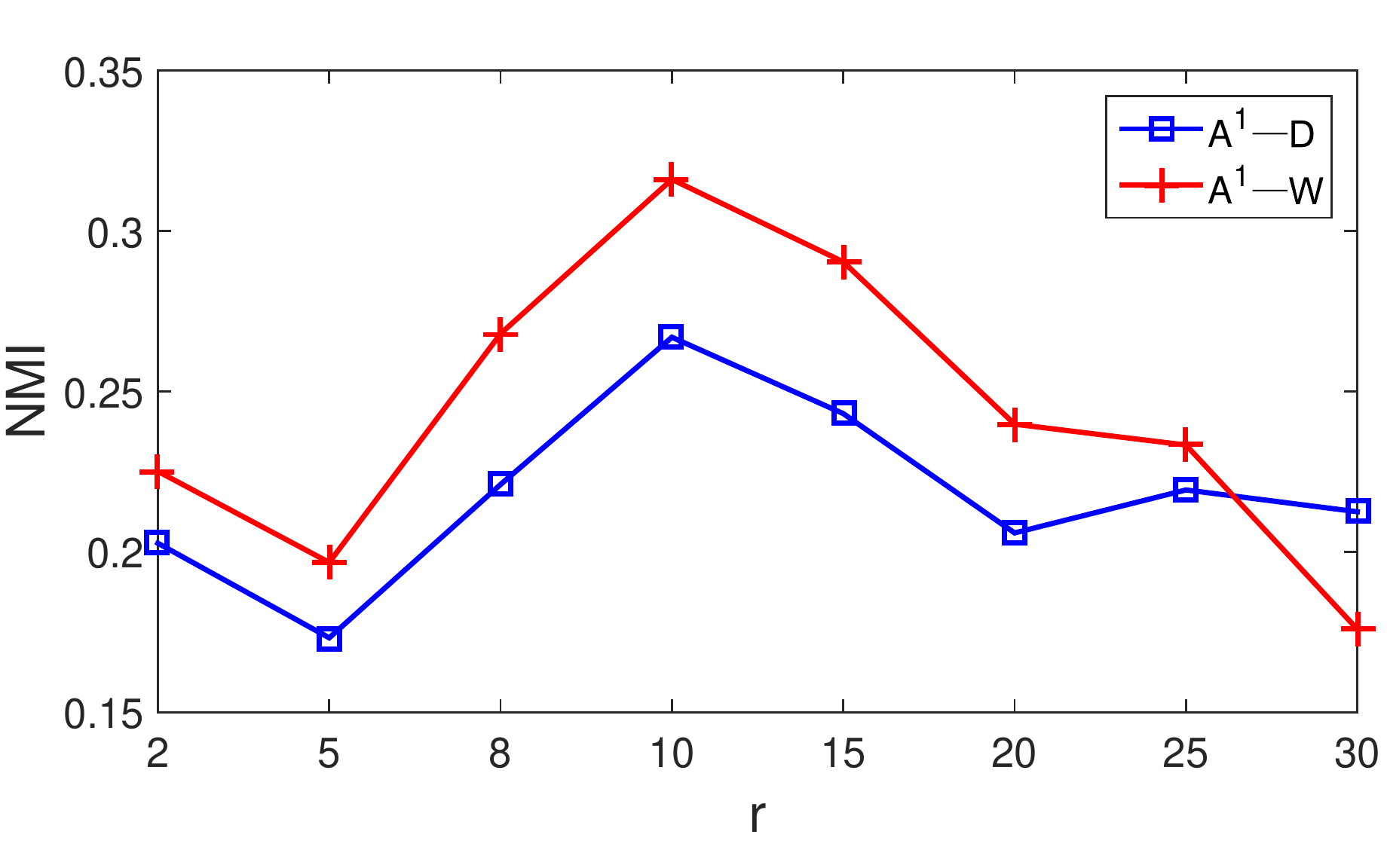}
  \end{minipage}
  }
  \caption{the sensitivity of hyperparameters in the Dslr and Webcam datasets. (a) and (b) represent the hyperparameter variation of the Dslr and Webcam between $\beta$ and $\gamma$ in A$^1 \rightarrow$D, W. (c) and (d) is the variation of $\eta$ and $r$
in A$^1 \rightarrow$D, W respectively}
  \label{Fig:7}
\end{figure*}
In Tables 1 and 2, we attain the performance of all compared methods. From these results, we can draw several conclusions. First, PA-1SmT is obviously superior to other methods, especially on the NMI index. Second, for single target domain, our method also performs well. It is worth noting that the relative improvement is nearly $9 \%$ in A$^1 \rightarrow$D. Third, with respect to the comparison of two and three target domains, the clustering performance is dramatically improved. Besides, for A$^1 \rightarrow$C$^2$ and C$^1\rightarrow$A$^2$, our method is slightly less than other methods in the knowledge transfer due to the weak correlation between the Amazon and Dslr sub-datasets. Such a reason weakens the ability using the common model parameter to adaptively represent the target model parameters. From Fig.6, we can see that the clustering performance will improve with the increase of shared category numbers. In especial, for A$^1 \rightarrow$D and C$^1 \rightarrow$D, when the shared category number increases from 4 to 6, the Dslr sub-dataset has the most improvement in the clustering performance. The main reason is that there are strong correlation between two sub-datasets, which provides a good common model parameter dictionary. Thereby, the target model parameters is well-adapted.

In addition, we use A$^1 \rightarrow$W, D as an example to analyze the sensitivity of hyperparameters. From Fig.7, we witness the impact of hyperparameter change on knowledge transfer. The hyperparameters $\beta$ and $\gamma$ respectively represent the transfer degree from the source domain to the individual target domains or among multiple target domains. In (a) and (b), the performance change performs on a set of varied $\beta=\gamma=\{0.0001,0.001,0.01,0.1,1,10,50,100\}$. Since training the target domains is joint and these target domains can get learning from each other, the hyperparameters $\beta$ and $\gamma$ need to trend to relatively optimal values in A$^1 \rightarrow$W and A$^1 \rightarrow$D simultaneously, i.e., $\beta=50,\ \gamma=1$. which means that the target domains get more knowledge from the source domain than other target domains. In (c), the hyperparameter $\eta=\{0.00001,0.0001,0.001,0.01,0.1,1,10\}$ controls the sparsity of the model parameters, i.e., when $\eta=0.1$, A$^1 \rightarrow$W and A$^1 \rightarrow$D attain the optimal clustering performance. Finally, for the model parameter dictionary size $r$, we evaluate the clustering performance based on different $r$ in $\{2,5,8,10,15,20,25,30\}$. Obviously, $r=10$ is optimal.

2)\emph {PIE dataset:} In the PIE dataset, we use the PIE05 and PIE27 sub-datasets with a relatively more samples as the source domain, other sub-datasets are the target domains. Furthermore, since the PIE dataset has 68 categories, we choose the top 20 categories for training the source data. For the target domains, we choose 9 categories and then share 4 categories between two target domains. Meanwhile, in order to verify the validity of the mentioned scenarios above, i.e., $ \bigcap_{j=1}^M {\cal C}_T^j =0 \land {\cal C}_T^j \cap {\cal C}_T^k={\cal C}_{c0}$ and $\bigcap_{j=1}^M {\cal C}_T^j ={\cal C}_c $, we train on three target domains. As an example, we adopt two representative scenarios to process the PIE29 sub-dataset. The first selects 9 categories and then shares 4 categories simultaneously with the other two target domains. The second selects 10 classes but shares five classes respectively with the other two target domains.

\begin{table}[ht]
\begin{center}
\footnotesize
\caption{Average NMI of All Compared Methods on PIE Datasets}
\label{Table 3}
\setlength{\tabcolsep}{1.5mm}{
\begin{tabular}{|c|c|c|c|c|c|c|c|}
\hline
\multirow{2}{*}{\begin{tabular}[c]{@{}c@{}}\\$D_S$\end{tabular}} & \multirow{2}{*}{\begin{tabular}[c]{@{}c@{}}\\$D_T$\end{tabular}} & baseline & \multicolumn{4}{c|}{1S1T}                                                        & 1SmT            \\ \cline{3-8}
                                                                         &                                                                          & SLMC     & TSC    & STC    & TFSC   & \begin{tabular}[c]{@{}c@{}}Single\\ ours\end{tabular} & ours            \\ \hline
\multirow{6}{*}{PIE05}                                                   & PIE07$^2$                                                                   & 0.1956   & \underline{\emph{0.4227}} & 0.4171 & 0.414  & 0.4115                                                & {\textbf{0.4316}} \\
                                                                         & PIE09$^1$                                                                   & 0.1839   & 0.4054 & 0.4038 & 0.4161 & \underline{\emph{0.4356}}                                                & {\textbf{0.4544}} \\ \cline{2-8}
                                                 & PIE07$^2$                                                                   & 0.1956   & \underline{\emph{0.4227}} & 0.4171 & 0.414  & 0.4115                                                & {\textbf{0.4275}} \\
                                                                         & PIE29$^1$                                                                   & 0.229   & 0.4096 & 0.4137 & 0.4165 & \underline{\emph{0.5066}}                                                & {\textbf{0.5436}} \\ \cline{2-8}
                                                & PIE09$^2$                                                                   & 0.1771   & 0.4423 & 0.4407 & 0.4287 & \underline{\emph{0.45}}                                                  & {\textbf{0.4864}} \\
                                                                         & PIE29$^1$                                                                   & 0.229   & 0.4096 & 0.4137 & 0.4165 & \underline{\emph{0.5066}}                                                & {\textbf{0.5415}} \\ \hline
\multirow{6}{*}{PIE27}                                                   & PIE07$^2$                                                                   & 0.1956   & 0.4214 & 0.4153 & 0.4083 & \underline{\emph{0.428}}                                                  & {\textbf{0.4493}} \\
                                                                         & PIE09$^1$                                                                   & 0.1839   & 0.4014 & 0.4041 & \underline{\emph{0.4094}} & 0.4033                                                & {\textbf{0.4224}} \\ \cline{2-8}
                                                  & PIE07$^2$                                                                   & 0.1956   & 0.4214 & 0.4153 & 0.4083 & \underline{\emph{0.428}}                                                  & {\textbf{0.4427}} \\
                                                                         & PIE29$^1$                                                                   & 0.229    & 0.4035 & 0.4061 & 0.4131 & \underline{\emph{0.4385}}                                                & {\textbf{0.5059}} \\ \cline{2-8}
                                                  & PIE09$^2$                                                                   & 0.1771   & \underline{\emph{0.4366}} & 0.4315 & 0.4136 & 0.4263                                                & {\textbf{0.4456}} \\
                                                                         & PIE29$^1$                                                                   & 0.229    & 0.4035 & 0.4061 & 0.4131 & \underline{\emph{0.4385}}                                                & {\textbf{0.498}}  \\ \hline
\end{tabular}}\\
\end{center}
\end{table}

\begin{table}[ht]
\begin{center}
\footnotesize
\caption{Average RI of All Compared Methods on PIE Datasets}
\label{Table 4}
\setlength{\tabcolsep}{1.5mm}{
\begin{tabular}{|c|c|c|c|c|c|c|c|}
\hline
\multirow{2}{*}{\begin{tabular}[c]{@{}c@{}}\\$D_S$\end{tabular}} & \multirow{2}{*}{\begin{tabular}[c]{@{}c@{}}\\$D_T$\end{tabular}} & baseline & \multicolumn{4}{c|}{1S1T}                                                                 & 1SmT            \\ \cline{3-8}
                                                                         &                                                                          & SLMC     & TSC    & STC    & TFSC            & \begin{tabular}[c]{@{}c@{}}Single\\ ours\end{tabular} & ours            \\ \hline
\multirow{6}{*}{PIE05}                                                   & PIE07$^2$                                                                   & 0.7982   & 0.8443 & 0.8429 & 0.8372          & \underline{\emph{0.8479}}                                                & {\textbf{0.8505}} \\
                                                                         & PIE09$^1$                                                                   & 0.7735   & 0.8361 & 0.8339 & 0.8441          & \underline{\emph{0.8513}}                                                & {\textbf{0.8517}} \\ \cline{2-8}
                                                   & PIE07$^2$                                                                   & 0.7982   & 0.8443 & 0.8429 & 0.8372          & \underline{\emph{0.8479}}                                                & {\textbf{0.8494}} \\
                                                                         & PIE29$^1$                                                                   & 0.8176   & 0.8436 & 0.8457 & 0.858           & \underline{\emph{0.8761}}                                                & {\textbf{0.8859}} \\ \cline{2-8}
                                                  & PIE09$^2$                                                                   & 0.8055   & 0.8389 & 0.8379 & 0.8474          & \underline{\emph{0.8569}}                                                & {\textbf{0.8699}} \\
                                                                         & PIE29$^1$                                                                   & 0.8176   & 0.8436 & 0.8457 & 0.858           & \underline{\emph{0.8761}}                                                & {\textbf{0.8804}} \\ \hline
\multirow{6}{*}{PIE27}                                                   & PIE07$^2$                                                                   & 0.7982   & 0.8397 & 0.8311 & 0.8222          & \underline{\emph{0.8467}}                                                & {\textbf{0.8529}} \\
                                                                         & PIE09$^1$                                                                   & 0.7735   & 0.8234 & 0.8236 & \underline{\emph{0.8386}}          & 0.8377                                                & {\textbf{0.8415}} \\ \cline{2-8}
                                                & PIE07$^2$                                                                   & 0.7982   & 0.8397 & 0.8311 & 0.8222          & \underline{\emph{0.8467}}                                                & {\textbf{0.8538}} \\
                                                                         & PIE29$^1$                                                                   & 0.8176   & 0.8442 & 0.843  & 0.8498          & \underline{\emph{0.8527}}                                                & {\textbf{0.8709}} \\ \cline{2-8}
                                                  & PIE09$^2$                                                                   & 0.8055   & 0.8259 & 0.8255 & \underline{\emph{0.8496}} & 0.8472                                                & {\textbf{0.8504}} \\
                                                                         & PIE29$^1$                                                                   & 0.8176   & 0.8442 & 0.843  & 0.8498          & \underline{\emph{0.8527}}                                                & {\textbf{0.8711}} \\ \hline
\end{tabular}}
\end{center}
\end{table}

From the results in Tables 3 and 4, we can draw several conclusions. First, PA-1SmT is completely better than other methods on NMI and RI, and is also superior to most others in a single domain. Second, when the PIE29 sub-dataset is added to the target domains, PA-1SmT can improve nearly $10\% \thicksim 14\%$ on NMI. This mainly owes to the connection of the common model parameter dictionary and its adaptive sparse representation, whereas TSC, STC and TFSC methods transfer knowledge without adaptive sparse representation, which leads to relatively poor clustering performance. Besides, for PIE05$ \rightarrow$PIE07 and PIE27$\rightarrow$PIE09 in a single domain, although our method is inferior to other methods on NMI, it is only lower 1\% than other methods. On the other hand, on RI, our method in PIE05$ \rightarrow$PIE07 is higher than others, and in PIE27$\rightarrow$PIE09 is almost equal to TFSC. In Table 5, we conduct experiments on three target domains to verify that the proposed method is able to handle the scenarios not only $ \bigcap_{j=1}^M {\cal C}_T^j =0 \land {\cal C}_T^j \cap {\cal C}_T^k={\cal C}_{c0}$ but also $\bigcap_{j=1}^M {\cal C}_T^j ={\cal C}_c $. Meanwhile, after adding a target domain, the clustering performance of the other two target domains has also improved.

\begin{table}[ht]
\begin{center}
\footnotesize
\caption{The Result of Two Representative Scenarios on Three Target Domains}
\label{Table 5}
\setlength{\tabcolsep}{1mm}{
\begin{tabular}{|c|c|c|c|c|c|c|}
\hline
                                               & \begin{tabular}[c]{@{}c@{}}$D_S$\end{tabular} & \begin{tabular}[c]{@{}c@{}}$D_T$\end{tabular} & NMI                         & \begin{tabular}[c]{@{}c@{}}Increme-\\ntal NMI\end{tabular} & RI                          & \begin{tabular}[c]{@{}c@{}}Increme-\\ntal NMI\end{tabular} \\ \hline
\multirow{6}{*}{\begin{tabular}[c]{@{}c@{}}$ \bigcap_{j=1}^M {\cal C}_T^j =0$ \\ $\land {\cal C}_T^j \cap {\cal C}_T^k={\cal C}_{c0}$\end{tabular}}                        & \multirow{3}{*}{PIE05}                                  & PIE07$^2$                                                  & 0.4904                      & \textbf{ 5.88 $\uparrow$}                                                     & 0.8636                      & \textbf{ 1.31 $\uparrow$ }                                                   \\
                                               &                                                         & PIE09$^1$                                                  & 0.4906                      & \textbf{3.62 $\uparrow$}                                                     & 0.8676                      & \textbf{1.59 $\uparrow$}                                                   \\
                                               &                                                         & PIE29$^3$                                                  & 0.5069                      & -                                                       & 0.878                       & -                                                      \\ \cline{2-7}
                                               & \multirow{3}{*}{PIE27}                                  & PIE07$^2$                                                  & 0.4697                      & \textbf{2.04 $\uparrow$}                                                     & 0.8524                      & 0.05 $\downarrow$                                                   \\
                                               &                                                         & PIE09$^1$                                                  & 0.4372                      & \textbf{1.48  $\uparrow$ }                                                   & 0.8532                      & \textbf{1.17   $\uparrow$}                                                  \\
                                               &                                                         & PIE29$^3$                                                  & 0.4556                      & -                                                         & 0.8641                      & -                                                        \\ \hline
\multicolumn{1}{|l|}{\multirow{6}{*}{ $\bigcap_{j=1}^M {\cal C}_T^j ={\cal C}_c $}} & \multirow{3}{*}{PIE05}                                  & \multicolumn{1}{l|}{PIE07$^2$}                             & \multicolumn{1}{l|}{0.4605} & \textbf{2.89 $\uparrow$}                                                     & \multicolumn{1}{l|}{0.8555} & \textbf{0.5 $\uparrow$}
\\
\multicolumn{1}{|l|}{}                         &                                                         & \multicolumn{1}{l|}{PIE09$^1$}                             & \multicolumn{1}{l|}{0.5248} & \textbf{7.04  $\uparrow$}                                                    & \multicolumn{1}{l|}{0.8775} & \textbf{2.58 $\uparrow$}
\\
\multicolumn{1}{|l|}{}                         &                                                         & \multicolumn{1}{l|}{PIE29$^4$}                             & \multicolumn{1}{l|}{0.437}  & -                                                         & \multicolumn{1}{l|}{0.8508} & -                                                        \\ \cline{2-7}
\multicolumn{1}{|l|}{}                         & \multirow{3}{*}{PIE27}                                  & \multicolumn{1}{l|}{PIE07$^2$}                             & \multicolumn{1}{l|}{0.4603} & \textbf{1.1  $\uparrow$}                                                     & \multicolumn{1}{l|}{0.8586} & \textbf{0.57  $\uparrow$}
\\
\multicolumn{1}{|l|}{}                         &                                                         & \multicolumn{1}{l|}{PIE09$^1$}                             & \multicolumn{1}{l|}{0.4529} & \textbf{ 3.05    $\uparrow$}                                                  & \multicolumn{1}{l|}{0.8486} & \textbf{0.71 }
 \\
\multicolumn{1}{|l|}{}                         &                                                         & \multicolumn{1}{l|}{PIE29$^4$}                             & \multicolumn{1}{l|}{0.4814} & -                                                         & \multicolumn{1}{l|}{0.866}  & -                                                        \\ \hline
\end{tabular}}\\
\end{center}
\footnotesize{ The $\uparrow$/$\downarrow$ represents the increase/decrease of performance.}
\end{table}

For the analysis of hyperparameters sensitivity in PIE datasets, we use PIE27$\rightarrow$PIE07, PIE29 as representative. It is similar to A$^1 \rightarrow$W, D on the range of hyperparameters $\beta,\ \gamma,\ \eta$ and $r$. From Fig.8 (a) and (b), when $\beta=50$ and $\gamma=10$, the clustering performance is relatively optimal, which shows that the target domains get more knowledge from the source domain than other target domains. In Fig. 8 (c) and (d), when $\eta=0.1$ and $r=15$, NMIs attain the optimal values respectively.

\begin{figure*}[ht]
  \centering
  \subfigure[]{
  \begin{minipage}{4cm}
  \centering
  \includegraphics[scale=0.22]{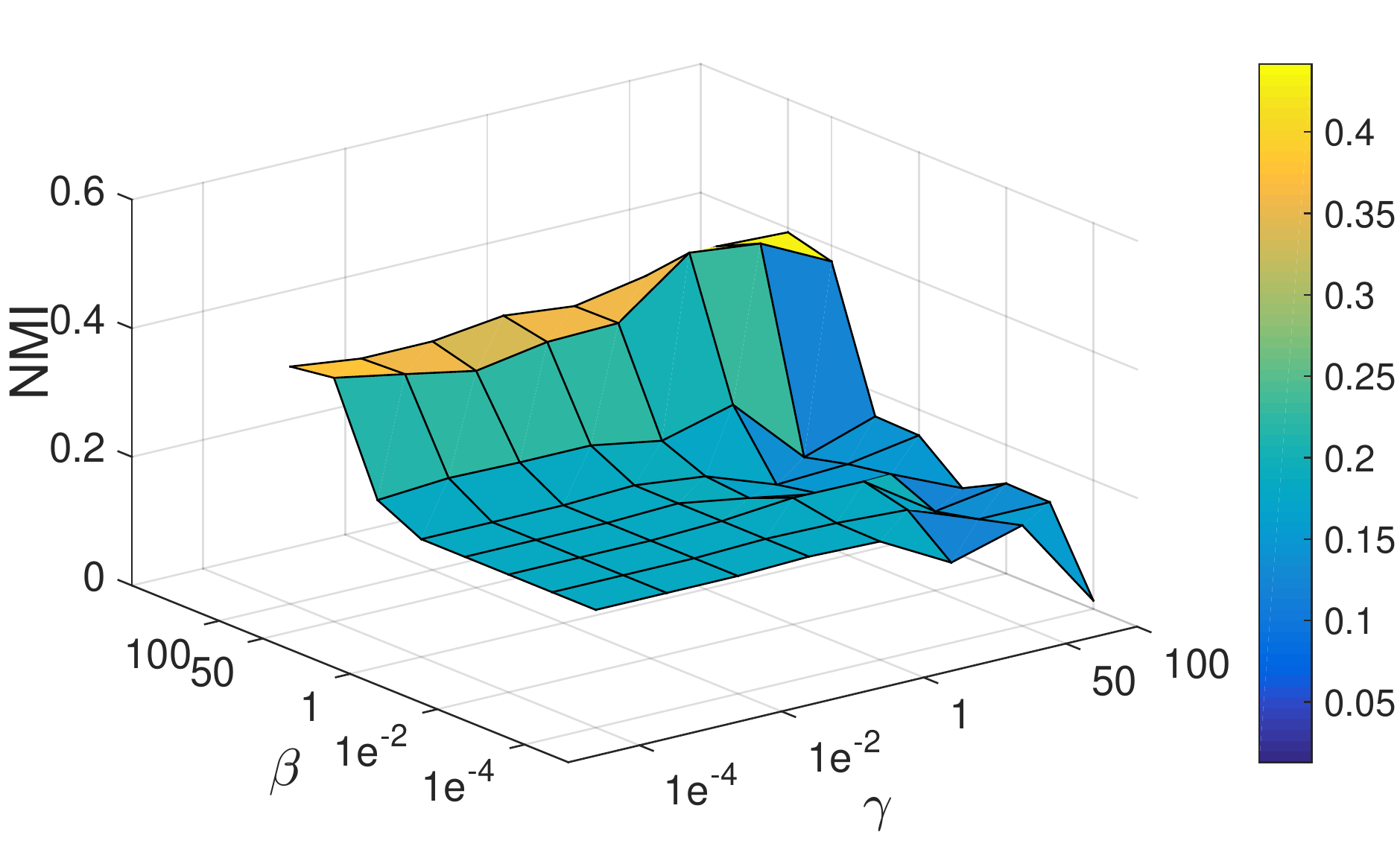}
  \end{minipage}
  }
  \subfigure[]{
  \begin{minipage}{4cm}
  \centering
  \includegraphics[scale=0.22]{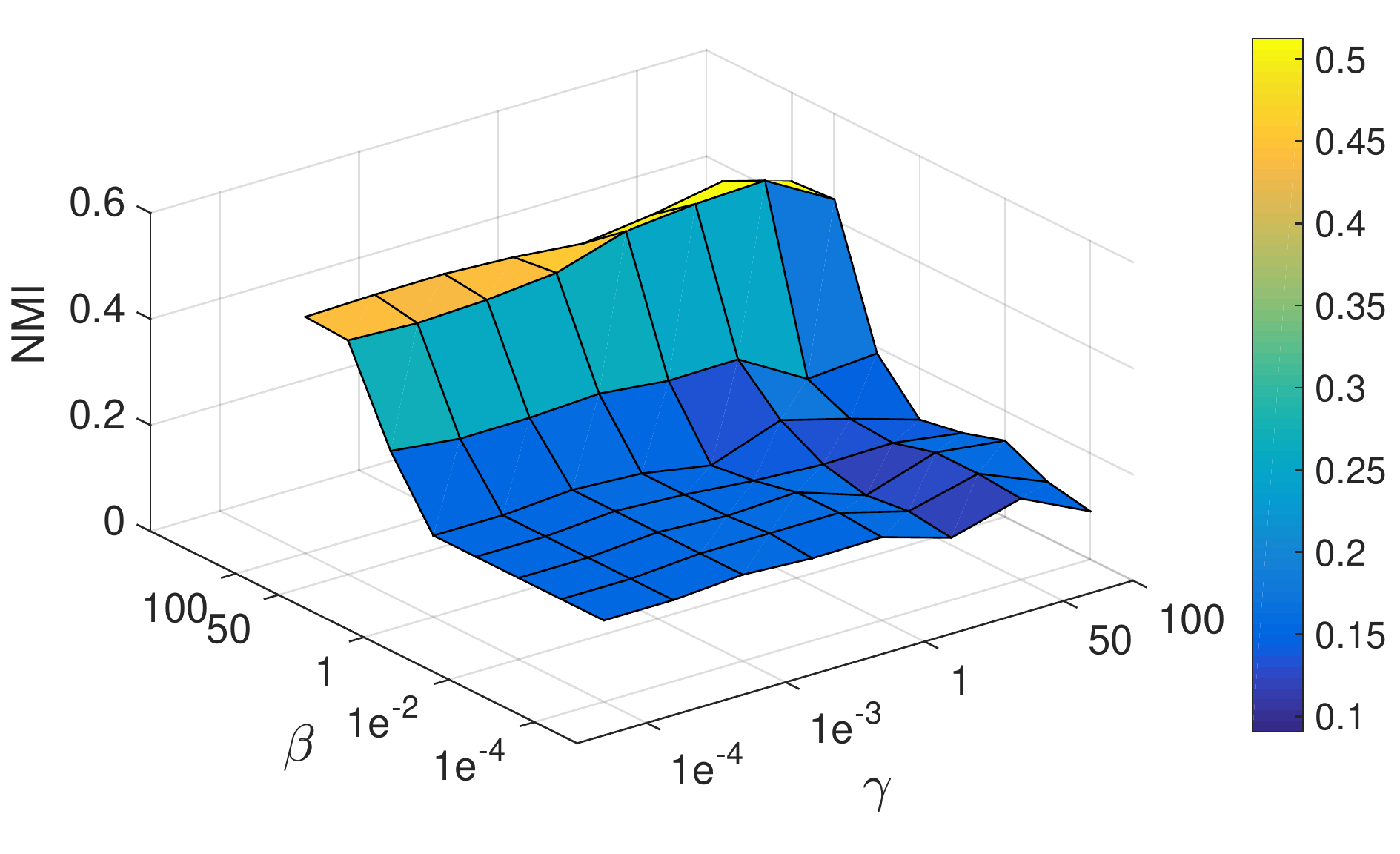}
  \end{minipage}
  }
  \subfigure[]{
  \begin{minipage}{4cm}
  \centering
  \includegraphics[scale=0.22]{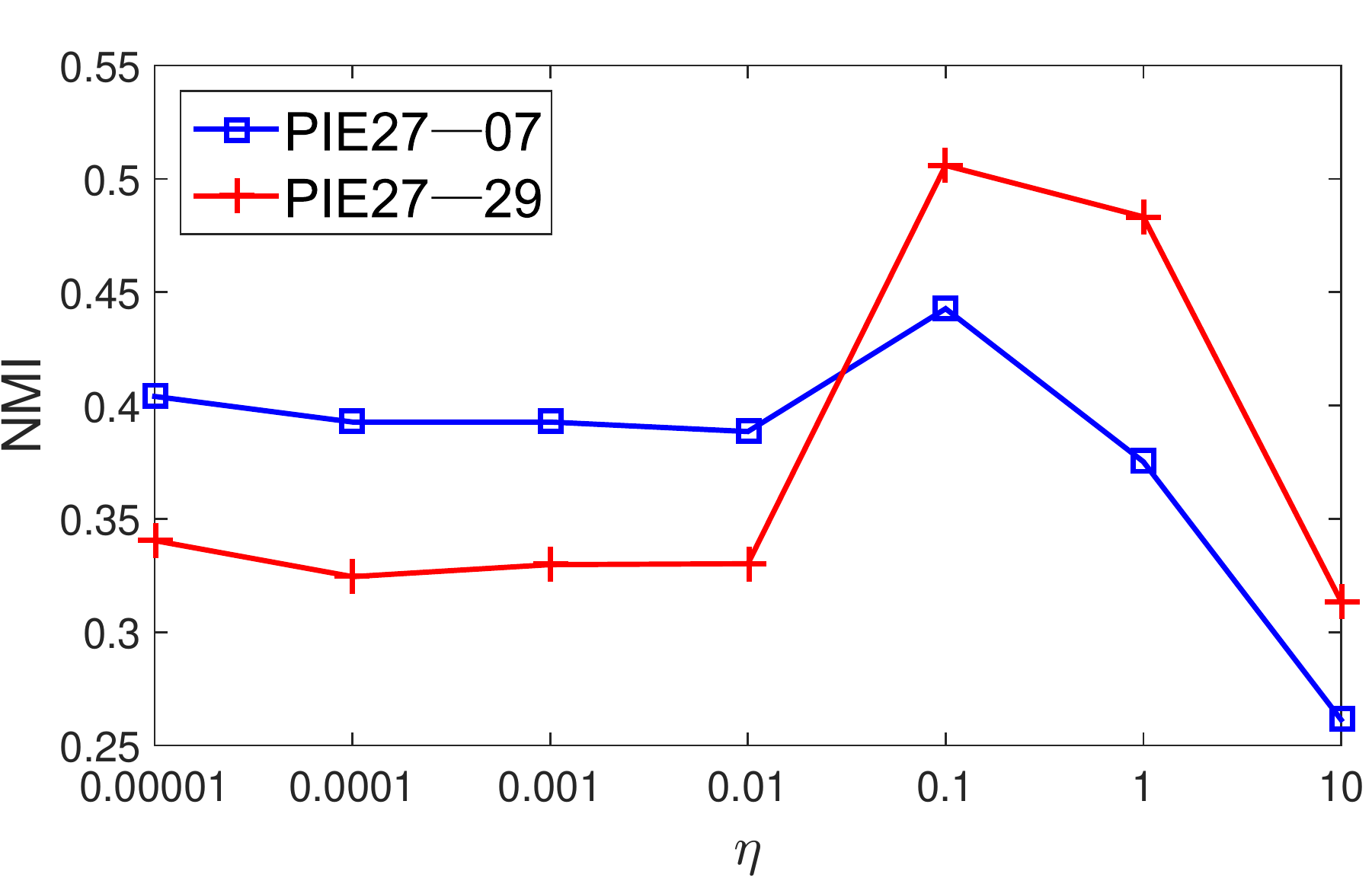}
  \end{minipage}
  }
  \subfigure[]{
  \begin{minipage}{4cm}
  \centering
  \includegraphics[scale=0.22]{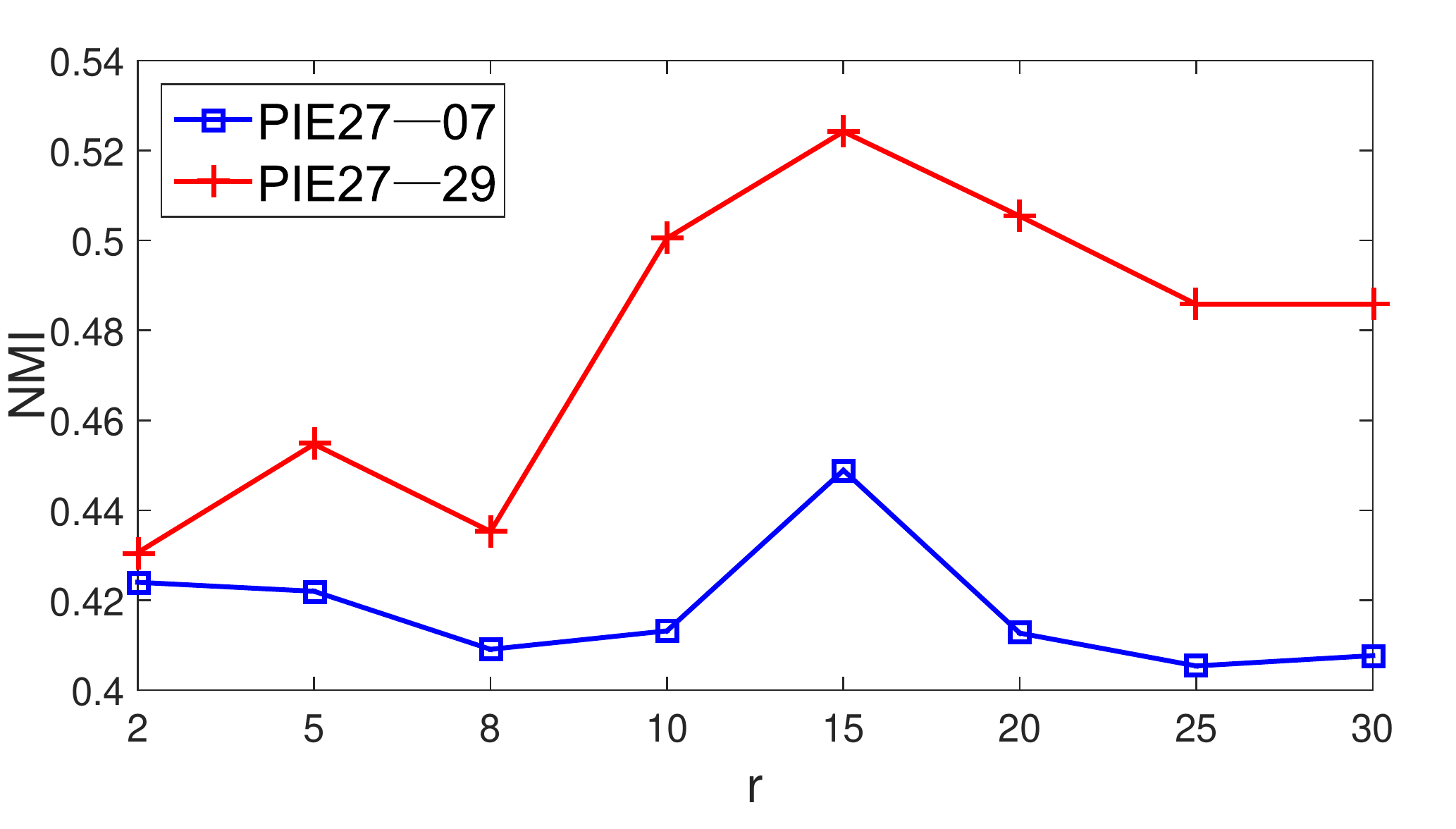}
  \end{minipage}
  }
  \caption{The sensitivity of hyperparameters in the PIE07 and PIE29 datasets. (a) and (b) represent the parameter variation of the PIE07 and PIE29 between $\beta$ and $\gamma$ in PIE27$\rightarrow$PIE07, PIE29. (c) and (d) are the variation of $\eta$ and $r$ in PIE27$\rightarrow$PIE07, PIE29 respectively}
  \label{Fig:8}
\end{figure*}

3)\emph {Extended Yale B dataset:} In the Yale B dataset, we experiment on three target domains, and respectively select Y2, Y3, Y4, Y5 as the source domain. In the source domain, we use all categories, i.e., 38 categories. In these target domains, there are in each domain. And 18 categories share 12 categories from the 7\emph{-th} to 18\emph{-th}. In addition, the Y1 sub-dataset is used as a newly added target domain, which shares the same 12 categories as other target domains.

\begin{table}[ht]
\begin{center}
\footnotesize
\caption{Average NMI of All Compared Methods on YaleB Datasets}
\label{my-label}
\setlength{\tabcolsep}{1.5mm}{
\begin{tabular}{|c|c|c|c|c|c|c|c|}
\hline
\multirow{2}{*}{\begin{tabular}[c]{@{}c@{}}\\$D_S$\end{tabular}} & \multirow{2}{*}{\begin{tabular}[c]{@{}c@{}}\\$D_T$\end{tabular}} & baseline & \multicolumn{4}{c|}{1S1T}                                                                         & 1SmT            \\ \cline{3-8}
                                                                         &                                                                          & SLMC     & TSC             & STC    & TFSC           & \begin{tabular}[c]{@{}c@{}}Single\\ ours\end{tabular} & ours            \\ \hline
\multirow{3}{*}{Y2}                                                      & Y5                                                                       & 0.3007   & {\textbf{0.3608}} & 0.3521 &\underline{\emph {0.3595}}   & 0.3392                                                & 0.3583          \\
                                                                         & Y3                                                                       & 0.3496   & 0.3829          & 0.3926 & 0.3912         & \underline{\emph{0.4377}}                                                & {\textbf{0.4621}} \\
                                                                         & Y4                                                                       & 0.3266   & 0.3926          & 0.3862 & 0.3964         & \underline{\emph{0.3973}}                                                & {\textbf{0.4108}} \\ \hline
\multirow{3}{*}{Y3}                                                      & Y5                                                                       & 0.3007   & {\textbf{0.3616}} & \underline{\emph{0.353}}  & 0.36           & 0.3416                                                & 0.3511          \\
                                                                         & Y2                                                                       & 0.368    & 0.4019          & 0.3912 & 0.3974         & \underline{\emph{0.4254}}                                                & {\textbf{0.4654}} \\
                                                                         & Y4                                                                       & 0.3266   & 0.3931          & 0.3878 & 0.3956         & \underline{\emph{0.3994}}                                                & {\textbf{0.4235}} \\ \hline
\multirow{3}{*}{Y4}                                                      & Y5                                                                       & 0.3007   & \underline{\emph{0.3589}}          & 0.3518 & {\textbf{0.363}} & 0.3335                                                & 0.3499          \\
                                                                         & Y2                                                                       & 0.368    & 0.3898          & 0.3944 & 0.3927         & \underline{\emph{0.4079}}                                                & {\textbf{0.4333}} \\
                                                                         & Y3                                                                       & 0.3496   & 0.3933          & 0.3903 & 0.4028         & \underline{\emph{0.4237}}                                                & {\textbf{0.4458}} \\ \hline
\multirow{3}{*}{Y5}                                                      & Y2                                                                       & 0.368   & 0.3808          & 0.3861 & 0.3944         & \underline{\emph{0.4156}}                                                & {\textbf{0.4416}} \\
                                                                         & Y3                                                                       & 0.3496   & 0.3863          & 0.3809 & 0.3927         & \underline{\emph{0.4198}}                                                & {\textbf{0.4462}}  \\
                                                                         & Y4                                                                       & 0.3266   & \underline{\emph{0.3928}} & 0.3844 & 0.3771         & 0.387                                                 & {\textbf{0.4094}}          \\ \hline
\end{tabular}}
\end{center}
\end{table}

\begin{table}[ht]
\begin{center}
\footnotesize
\caption{Average RI of All Compared Methods on YaleB Datasets}
\label{my-label}
\setlength{\tabcolsep}{1.5mm}{
\begin{tabular}{|c|c|c|c|c|c|c|c|}
\hline
\multirow{2}{*}{\begin{tabular}[c]{@{}c@{}}\\$D_S$\end{tabular}} & \multirow{2}{*}{\begin{tabular}[c]{@{}c@{}}\\$D_T$\end{tabular}} & baseline & \multicolumn{4}{c|}{1S1T}                                                                 & 1SmT            \\ \cline{3-8}
                                                                         &                                                                          & SLMC     & TSC             & STC    & TFSC   & \begin{tabular}[c]{@{}c@{}}Single\\ ours\end{tabular} & ours            \\ \hline
\multirow{3}{*}{Y2}                                                      & Y5                                                                       & 0.8697   & {\textbf{0.9168}} & \underline{\emph{0.9159}} & 0.9003 & 0.8976                                                & 0.9032          \\
                                                                         & Y3                                                                       & 0.8733   & 0.9152          & 0.9161 & 0.9062 & \underline{\emph{0.9176}}                                                & {\textbf{0.9236}} \\
                                                                         & Y4                                                                       & 0.8717   & {\textbf{0.9171}} & 0.913  & 0.9087 & 0.907                                                 & \underline{\emph{0.9169}}          \\ \hline
\multirow{3}{*}{Y3}                                                      & Y5                                                                       & 0.8697   & {\textbf{0.9166}} & \underline{\emph{0.9164}} & 0.9014 & 0.9024                                                & 0.9035          \\
                                                                         & Y2                                                                       & 0.8764   & 0.9164          & 0.916  & 0.9088 & \underline{\emph{0.9158}}                                                & {\textbf{0.9212}} \\
                                                                         & Y4                                                                       & 0.8717   & \underline{\emph{0.9178}} & 0.9172 & 0.9071 & 0.9069                                                & {\textbf{0.9181}}          \\ \hline
\multirow{3}{*}{Y4}                                                      & Y5                                                                       & 0.8697   & {\textbf{0.9166}} & \underline{\emph{0.9169}} & 0.9025 & 0.9015                                                & 0.9044          \\
                                                                         & Y2                                                                       & 0.8764   & 0.9163          & 0.9158 & 0.9017 & \underline{\emph{0.9169}}                                                & {\textbf{0.9208}} \\
                                                                         & Y3                                                                       & 0.8733   & 0.9154          & 0.9155 & 0.9101 & \underline{\emph{0.9165} }                                               & {\textbf{0.9224}} \\ \hline
\multirow{3}{*}{Y5}                                                      & Y2                                                                       & 0.8764   & \underline{\emph{0.9154}}          & 0.9132 & 0.9063 & 0.9138                                                & {\textbf{0.9207}} \\
                                                                         & Y3                                                                       & 0.8733    & \underline{\emph{0.9159}} & 0.9155 & 0.905  & 0.9133                                                & {\textbf{0.9215}}          \\
                                                                         & Y4                                                                       & 0.8717   & {\textbf{0.917}}  & 0.9125 & 0.9028 & 0.9026                                                & \underline{\emph{0.9164 }}         \\ \hline
\end{tabular}}
\end{center}
\end{table}

\begin{figure}[ht]
\begin{center}
  \includegraphics[scale=0.4]{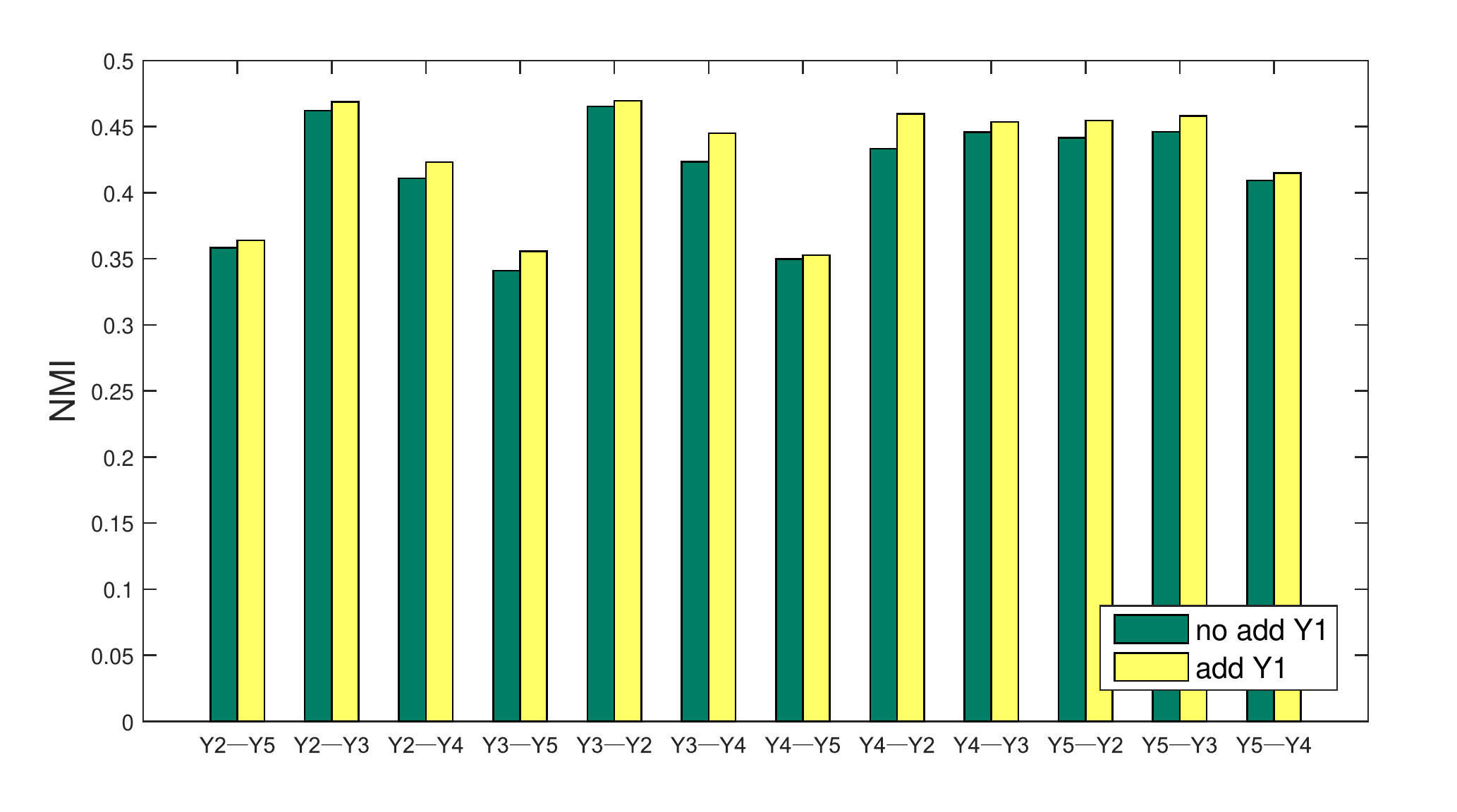}
  \caption{The comparison after adding a new target domain. The first line represents NMI without adding Y1 dataset, the second line represents NMI with adding Y1 dataset.}
  \label{Fig:9}
\end{center}
\end{figure}
From the results in Tables 6 and 7, we draw several conclusions. First, our method is mostly better than other methods not only in the 1SmT but also 1S1T. Second, when the Y2 and Y3 datasets are used as the source domain respectively, PA-1SmT can be increased by up to 7\% on NMI. Third, although Y4 and Y5 bear severe light variations, our method can also improve nearly by 2\% for the target domains Y2 and Y3, which means PA-1SmT can well transfer the knowledge from one source domain in the context of strong interference, mainly owing to the adaptive sparse representation of common model parameter dictionary. Meanwhile, the comparison result in RI is basically the same that in NMI. In addition, from Fig.9, we obviously find that the clustering performance has get increase after adding the Y1 dataset, the reason is adding a new target domain can provide more transferable knowledge and enrich the common model parameter dictionary, which helps not only other target domains but also itself.

\begin{figure*}[ht]
  \centering
  \subfigure[]{
  \begin{minipage}{5cm}
  \centering
  \includegraphics[scale=0.22]{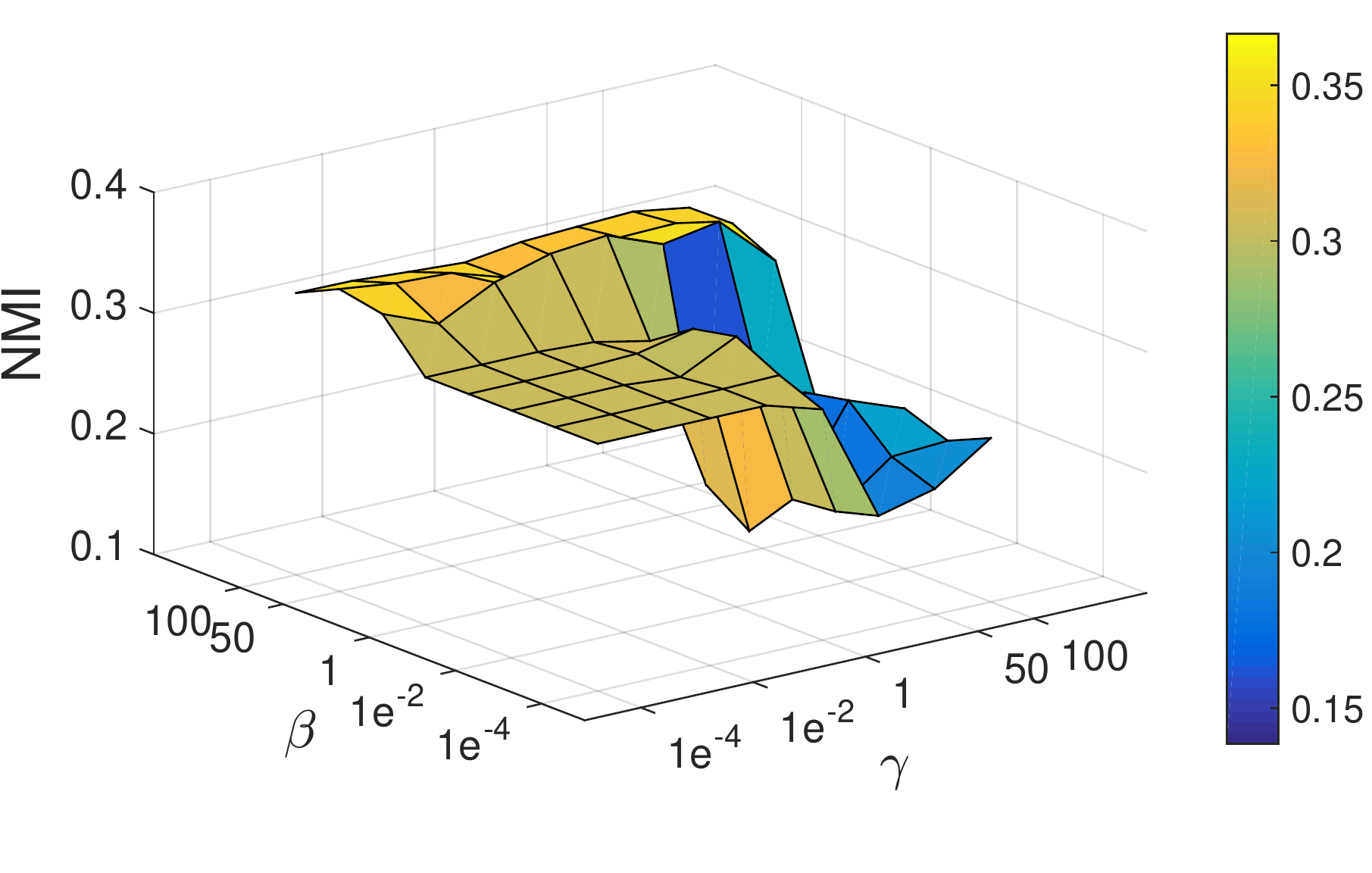}
  \end{minipage}
  }
  \subfigure[]{
  \begin{minipage}{5cm}
  \centering
  \includegraphics[scale=0.22]{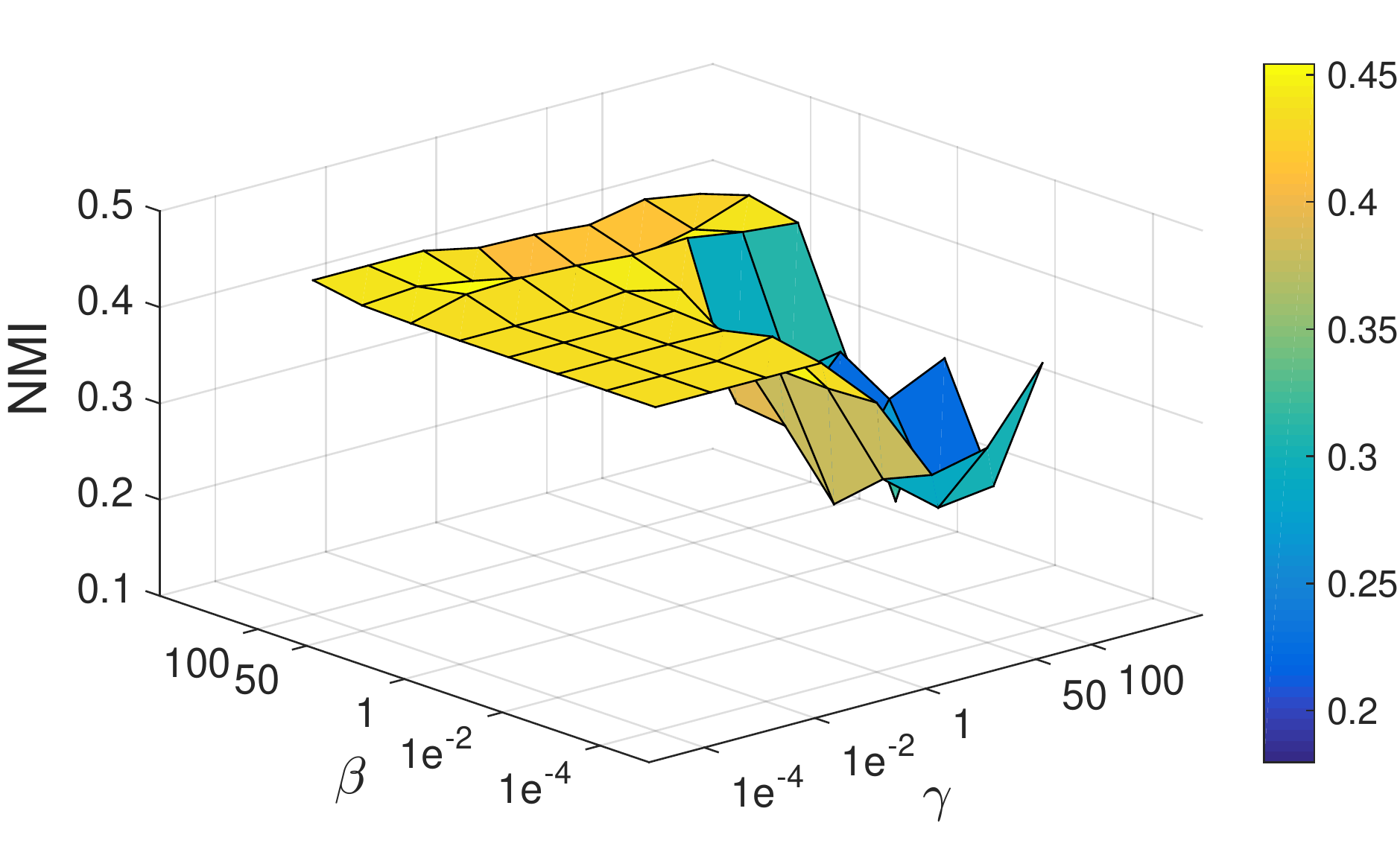}
  \end{minipage}
  }
  \subfigure[]{
  \begin{minipage}{5cm}
  \centering
  \includegraphics[scale=0.22]{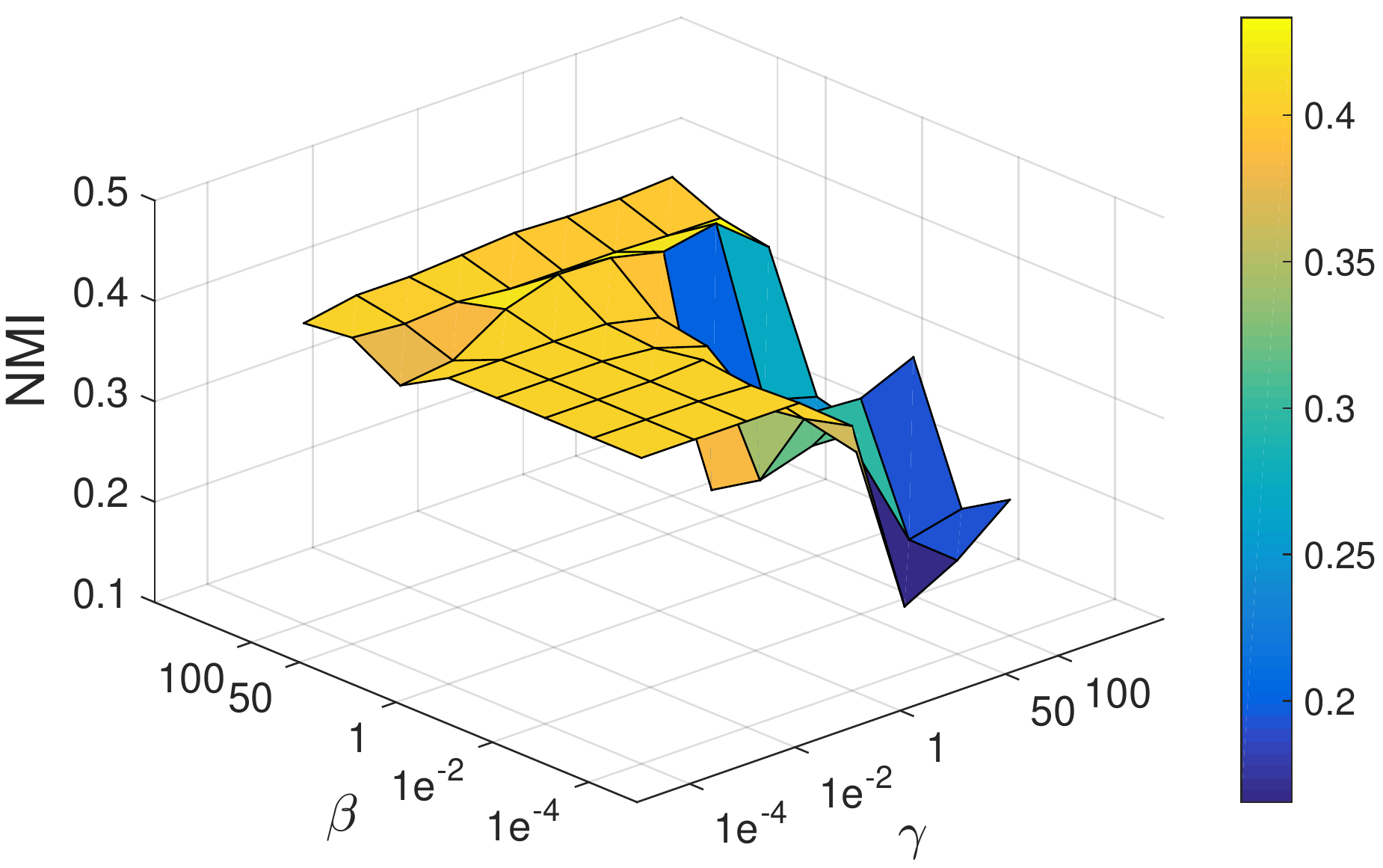}
  \end{minipage}
  }
  \subfigure[]{
  \begin{minipage}{7cm}
  \centering
  \includegraphics[scale=0.22]{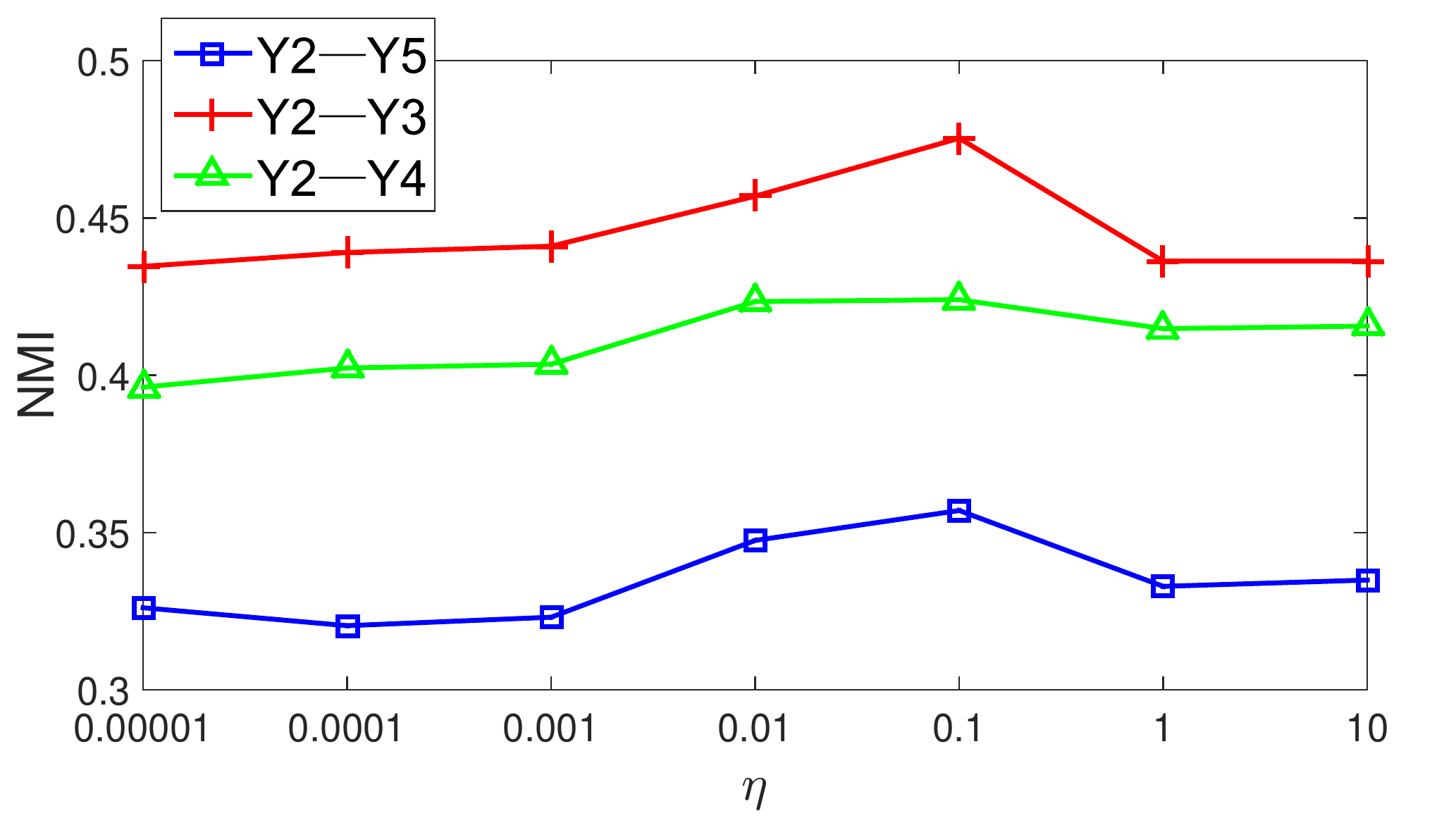}
  \end{minipage}
  }
  \subfigure[]{
  \begin{minipage}{7cm}
  \centering
  \includegraphics[scale=0.22]{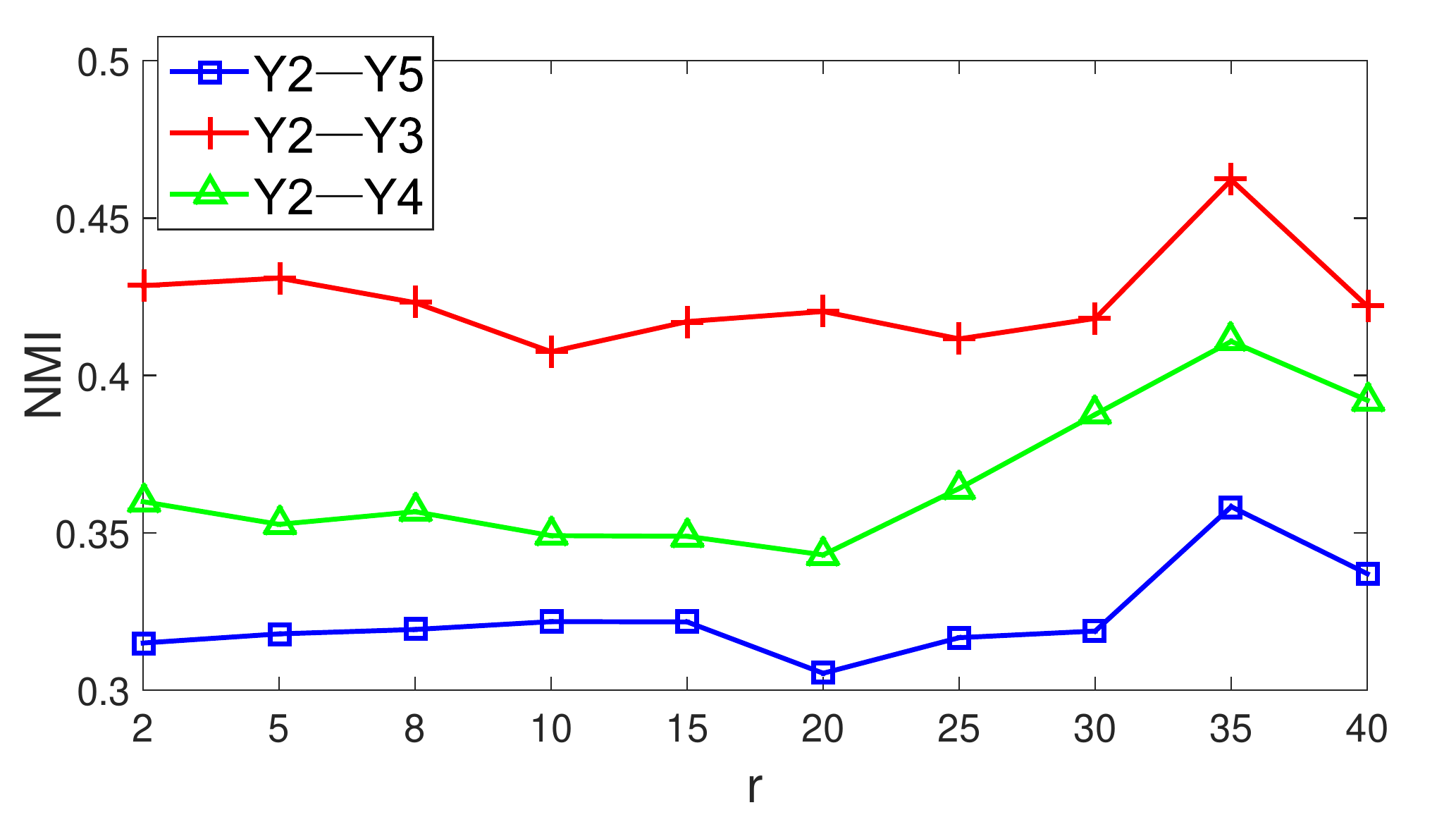}
  \end{minipage}
  }
  \caption{The sensitivity of hyperparameters in the Y5, Y3 and Y4 datasets. (a) ,(b) and (c) represent the hyperparameter variation of the Y5, Y3 and Y4 between $\beta$ and $\gamma$ in Y2$\rightarrow$Y5, Y3, Y4. (d) and (f) are the variation of $\eta$ and $r$ in Y2$\rightarrow$Y5, Y3, Y4 respectively}
  \label{Fig:10}
\end{figure*}

In the YaleB dataset, we choose Y2$\rightarrow$Y5, Y3, Y4 as an example to analyze the hyperparameters sensitivity. The range setting of hyperparameters $\beta,\ \gamma,\ \eta$ is consistent with the setting above. However, there are 38 categories in the source domain, thus $r$ is searched in $\{2,5,8,10,15,20,25,30,35,40\}$. From Fig.10 (a)-(c), the clustering performance achieves optimal when $\beta=10$ and $\gamma=0.1$, which means that the target domain acquires more knowledge from the source domain than from other target domains. In Fig.10 (d) and (f), when $\eta=0.1$ and r=35, Y2$ \rightarrow$Y5, Y2$\rightarrow$Y3 and Y2$\rightarrow$Y4 simultaneously attain the optimal clustering performance.

\subsection{Empirical Analysis}
All the experiments are performed on the computer with following specifications: Intel(R) Core(TM) i5-3470 CPU with PyCharm 2016. we use PIE27$\rightarrow$PIE07, PIE29 as an example to analyze the convergence of PA-1SmT, which is shown in Fig.11. We can observe that the objective function value decreases monotonically with the increase of the iterative number, which demonstrates that the iterative solving process of PA-1SmT is indeed convergent. Moreover, the NMI trends to increase with the growth of the iterative number, and the final result trends to stabilization.
\begin{figure}[ht]
  \centering
  \subfigure[]{
  \begin{minipage}{4cm}
  \centering
  \includegraphics[scale=0.22]{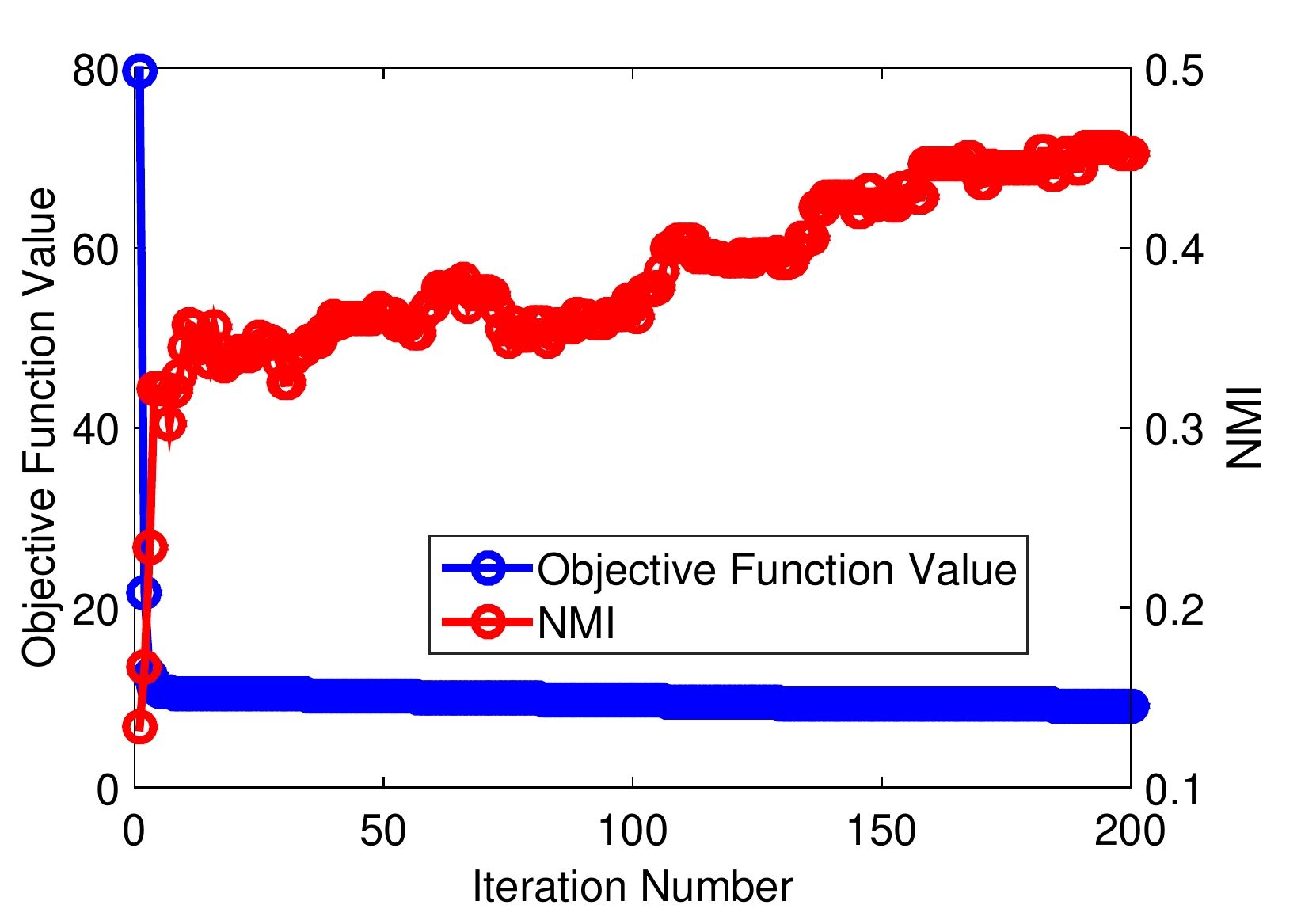}
  \end{minipage}
  }
  \subfigure[]{
  \begin{minipage}{4cm}
  \centering
  \includegraphics[scale=0.22]{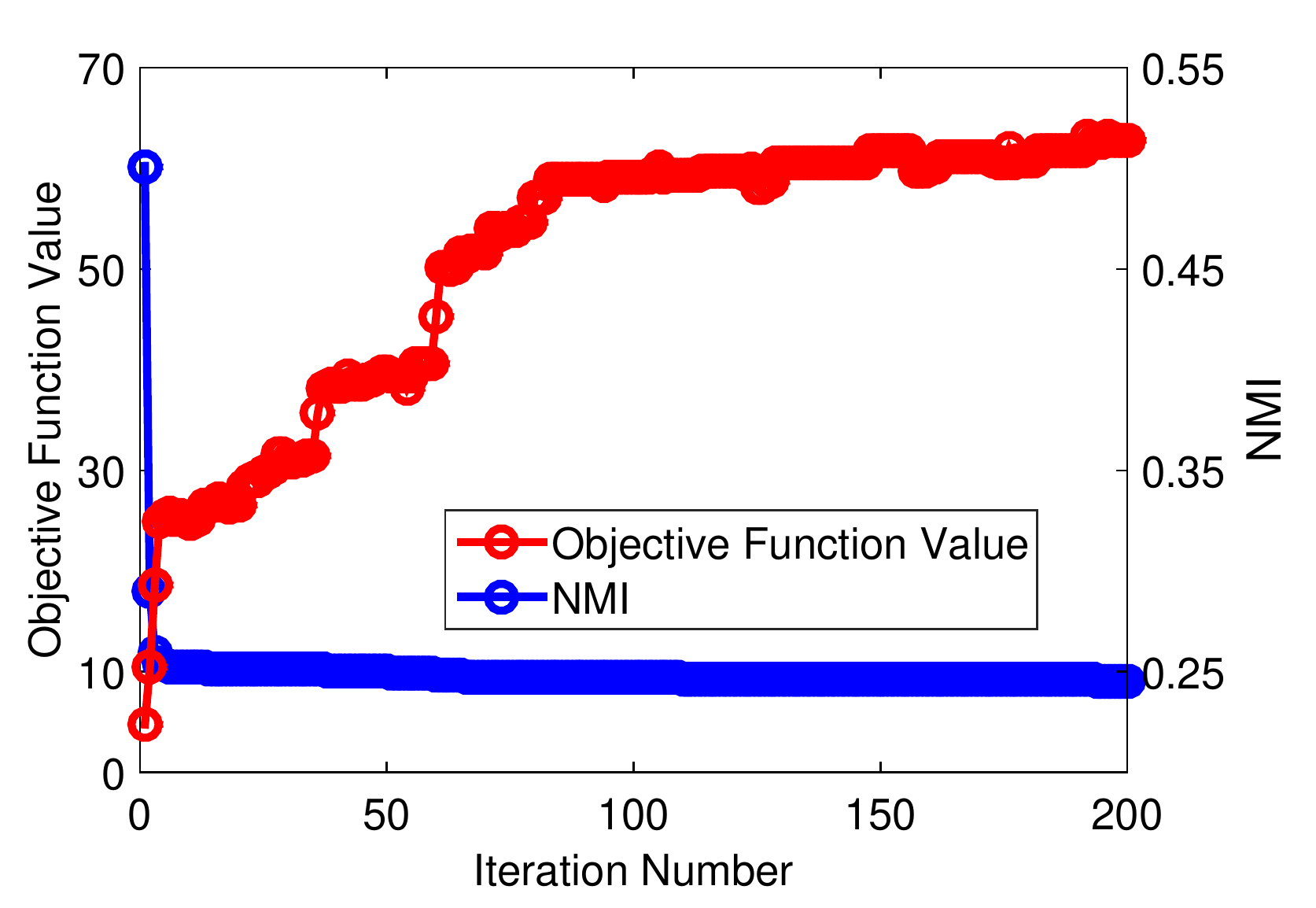}
  \end{minipage}
  }
  \caption{Convergence behavior of PA-1SmT on PIE27$\rightarrow$PIE07, PIE29. (a) and (b) represent the objective function value and NMI respectively on the PIE07 and PIE29 datasets.}
  \label{Fig:11}
\end{figure}

\section{Conclusion}
In this paper, we develop a new unsupervised domain adaptation scenario 1SmT. In such a new scenario, these unlabeled target domains may not necessarily share the exact same categories. Thus, we propose a novel model parameter adaptation framework (PA-1SmT) to solve this problem. Our approach constructs a common model parameter dictionary not only from the source domain to the individual target domains but also among these multiple target domains. Then we use it to sparsely represent individual target model parameters, which adaptively attains knowledge transfer among these domains. On the other hand, our approach transfers the knowledge based on the model parameters rather than data itself, which makes this approach available for DA of privacy protection. Finally, we use the alternating iterative strategy to optimize PA-1SmT, in which each step has a closed-form solution and can theoretically be guaranteed convergence. In the experiments, we conduct on three real datasets, which demonstrates the effectiveness of PA-1SmT compared with the state of the art, as well as improving the performance after adding a newly related target domain.

In the future, there are some worth-studying issues summarized as follows:

\hangafter 1 
\hangindent 2.6em 
$\bullet$ In this paper, our proposed 1SmT combined with mS1T can easily be extended to mSmT scenario, which provides more transferable knowledge among multiple domains. Meanwhile, the idea of model parameter adaptation can also apply to mS1T scenario by some model modification. Thus, in next work, we will adopt the neural network algorithm to solve such two scenarios.

\hangafter 1 
\hangindent 2.6em 
$\bullet$ Besides, we will research that scenario where the target categories are more than the source ones.


%

\ifCLASSOPTIONcompsoc
  \section*{}

\else
  \section*{}
\fi

\ifCLASSOPTIONcompsoc
  \section*{Acknowledgments}

\else
  \section*{Acknowledgment}
\fi

This work is supported by the National Natural Science Foundation of China (NSFC) under the Grant Nos. 61672281 and 61472186, the Key Program of NSFC under Grant No. 61732006.  Songcan Chen is the corresponding author.



%

%

\begin{IEEEbiography}
[{\includegraphics[width=1in,height=1.25in,clip,keepaspectratio]{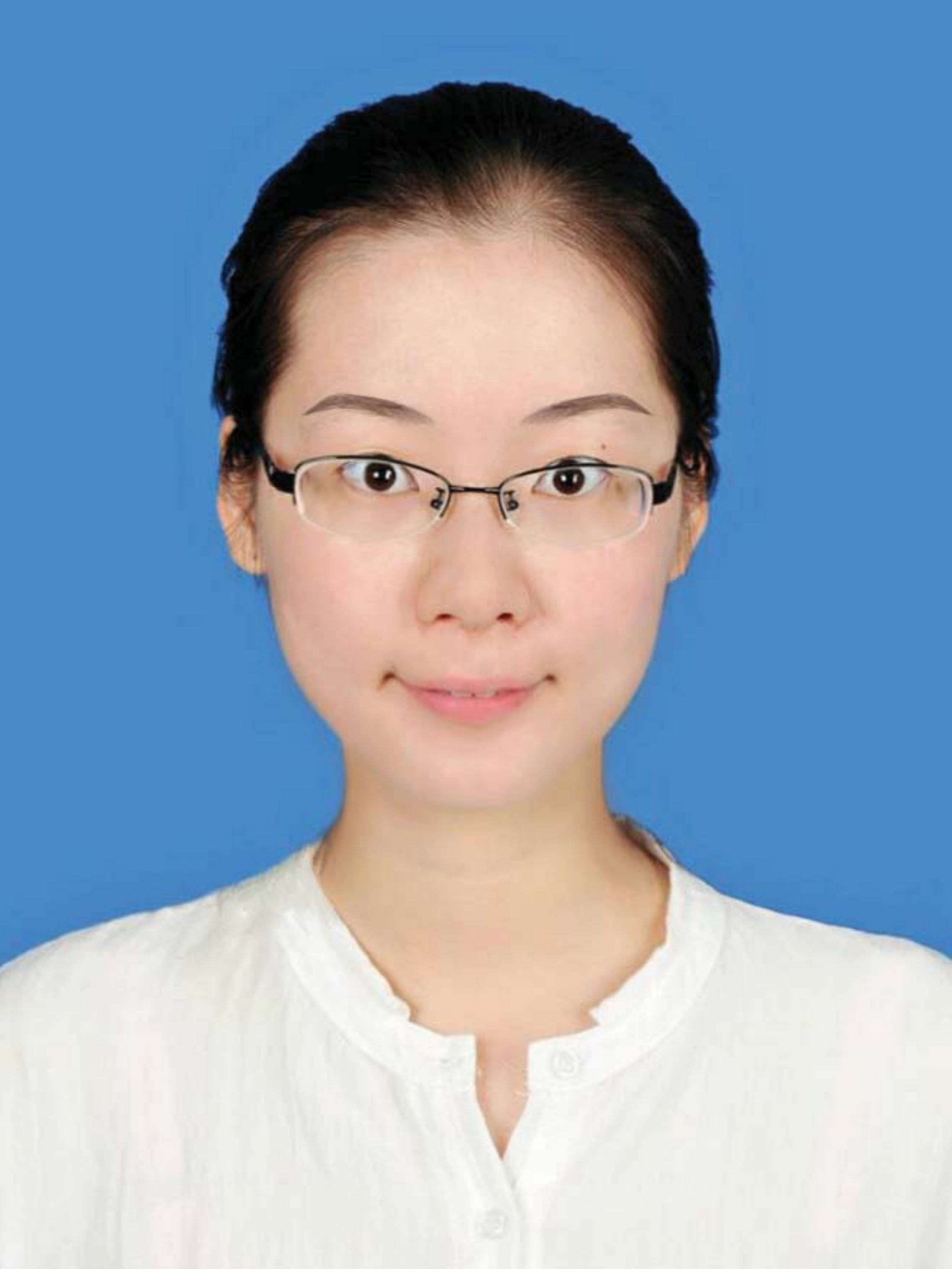}}]{Huanhuan Yu}
   received his B.S. degree in Information and Computing Science from Anhui University of Science and Technology in 2012 and 2016. She is currently senior M.S. graduate with the College of Computer Science \& Technology (CCST), Nanjing University of Aeronautics and Astronautics (NUAA). Her research interests include transfer learning, pattern recognition and machine learning.
\end{IEEEbiography}

\begin{IEEEbiography}
[{\includegraphics[width=1in,height=1.25in,clip,keepaspectratio]{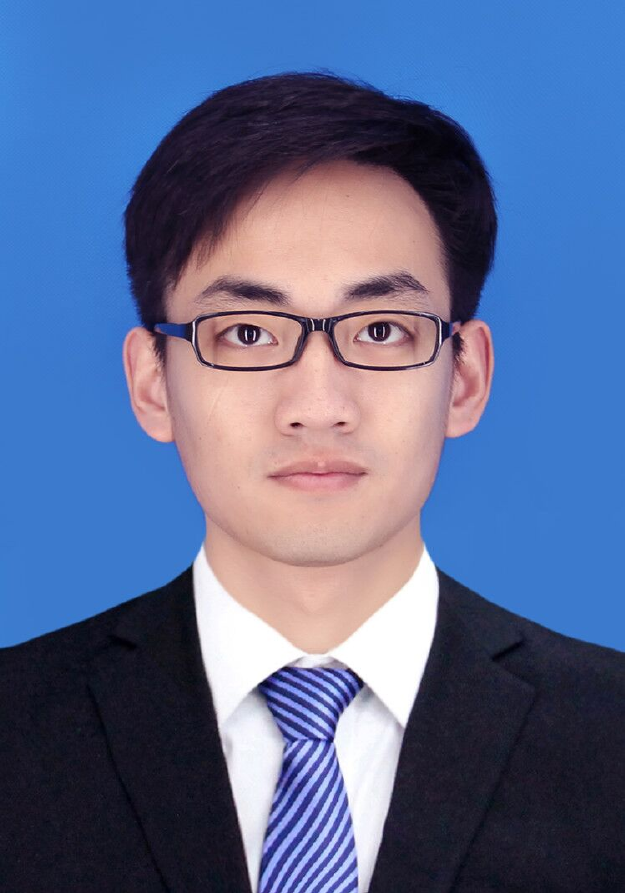}}]{Menglei Hu}
   received his B.S. degree in Information and Computing Science from Nanjing University of Aeronautics and Astronautics in 2012 and 2016. He is currently senior M.S. graduate with CCST, NUAA and has published a paper at IJCAI2018. His research interests include pattern recognition and machine learning.
\end{IEEEbiography}

\begin{IEEEbiography}
[{\includegraphics[width=1in,height=1.25in,clip,keepaspectratio]{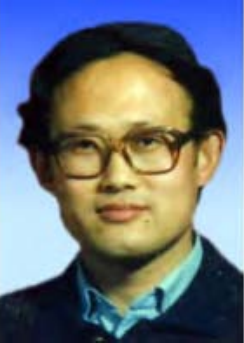}}]{Songcan Chen}
   received his B.S. degree in mathematics from Hangzhou University (now merged into Zhejiang University) in 1983. In 1985, he completed his M.S. degree in computer applications at Shanghai Jiaotong University and then worked at NUAA in January 1986. There he received a Ph.D. degree in communication and information systems in 1997. Since 1998, as a full-time professor, he has been with the College of Computer Science \& Technology at NUAA. His research interests include pattern recognition, machine learning and neural computing. He is also an IAPR Fellow.
\end{IEEEbiography}




\end{document}